\documentclass[11pt]{article}

\usepackage[preprint]{acl}

\usepackage{times}
\usepackage{latexsym}
\usepackage{bm}

\usepackage[T1]{fontenc}
\usepackage{upquote}
\usepackage[utf8]{inputenc}

\usepackage{microtype}

\usepackage{inconsolata}

\usepackage{graphicx}

\graphicspath{{latex/figs/}{figs/}}

\title{Self-Evolving Agents Can Repair Failures but Struggle to \\ Generalize from Them}
\newcommand*\samethanks[1][\value{footnote}]{\footnotemark[#1]}
\author{Weixiang Zhao$^1$\thanks{Equal contribution}, Yichen Zhang$^{1,2}$\samethanks, Yanyan Zhao$^1$, Jinming Bai$^1$, \\
        \textbf{Bin Liang}$^3$, \textbf{Ruifeng Xu}$^2$, \textbf{Bing Qin}$^1$, \textbf{Ting Liu}$^1$ \\
        $^1$Harbin Institute of Technology, Harbin, China \\
        $^2$Harbin Institute of Technology, Shenzhen, China \\
        $^3$The Chinese University of Hong Kong \\
        \texttt{\{wxzhao, yiczhang\}@ir.hit.edu.cn}}

\usepackage{booktabs}
\usepackage{colortbl}
\usepackage{array}
\usepackage{longtable}
\usepackage{caption}
\usepackage[most]{tcolorbox}
\definecolor{feedbackblue}{HTML}{EAF2F8}
\definecolor{feedbackrule}{HTML}{9CB3C8}
\newtcolorbox{feedbackcard}[1]{
  enhanced,
  colback=white,
  colframe=feedbackrule,
  colbacktitle=feedbackblue,
  coltitle=black,
  boxrule=0.7pt,
  arc=1.5mm,
  left=3mm,
  right=3mm,
  top=2.5mm,
  bottom=2.5mm,
  before skip=5mm,
  after skip=0pt,
  fonttitle=\bfseries\large,
  fontupper=\normalsize,
  title={#1}
}

\newcommand{\wx}[1]{\textcolor{black}{#1}}

\begin{document}
\maketitle

\begin{abstract}
Experience-driven self-evolution offers a promising path for large language model (LLM) agents to continually improve through interaction. Yet existing approaches often rely on successful trajectories or external evaluation signals, leaving unclear whether agents can evolve when such positive anchors are unavailable. We study this question through \emph{Cold-Start Failure-Only Evolution} (CFOE), where agents start without accumulated experience and learn only from failed interactions without privileged guidance. Across three self-evolving paradigms, multiple LLM backbones, and four interactive environments, we uncover a consistent \emph{repair--generalization gap}: failure-derived experience can help repair previously failed tasks, but rarely yields sustained improvements on unseen tasks. Further analysis shows that agents often misinterpret failures, while even verified-correct experience remains difficult to transfer. Relaxing CFOE reveals that verifier feedback and additional interaction provide limited gains, whereas successful trajectories substantially improve transfer. These findings suggest that current self-evolving agents are better at repairing observed failures than building generalizable capabilities from failure.
\end{abstract}

\section{Introduction}

With the arrival of the era of experience, large language model (LLM) agents are increasingly expected to improve through their own interactions \citep{silver2025welcome}. Self-evolving agents have therefore emerged as a promising paradigm that continually accumulates and reuses interaction experience \citep{dou2025evalearn,cai2025building,bell2025future,gao2025survey}.

\begin{figure}[h]
  \centering
  \includegraphics[width=\columnwidth]{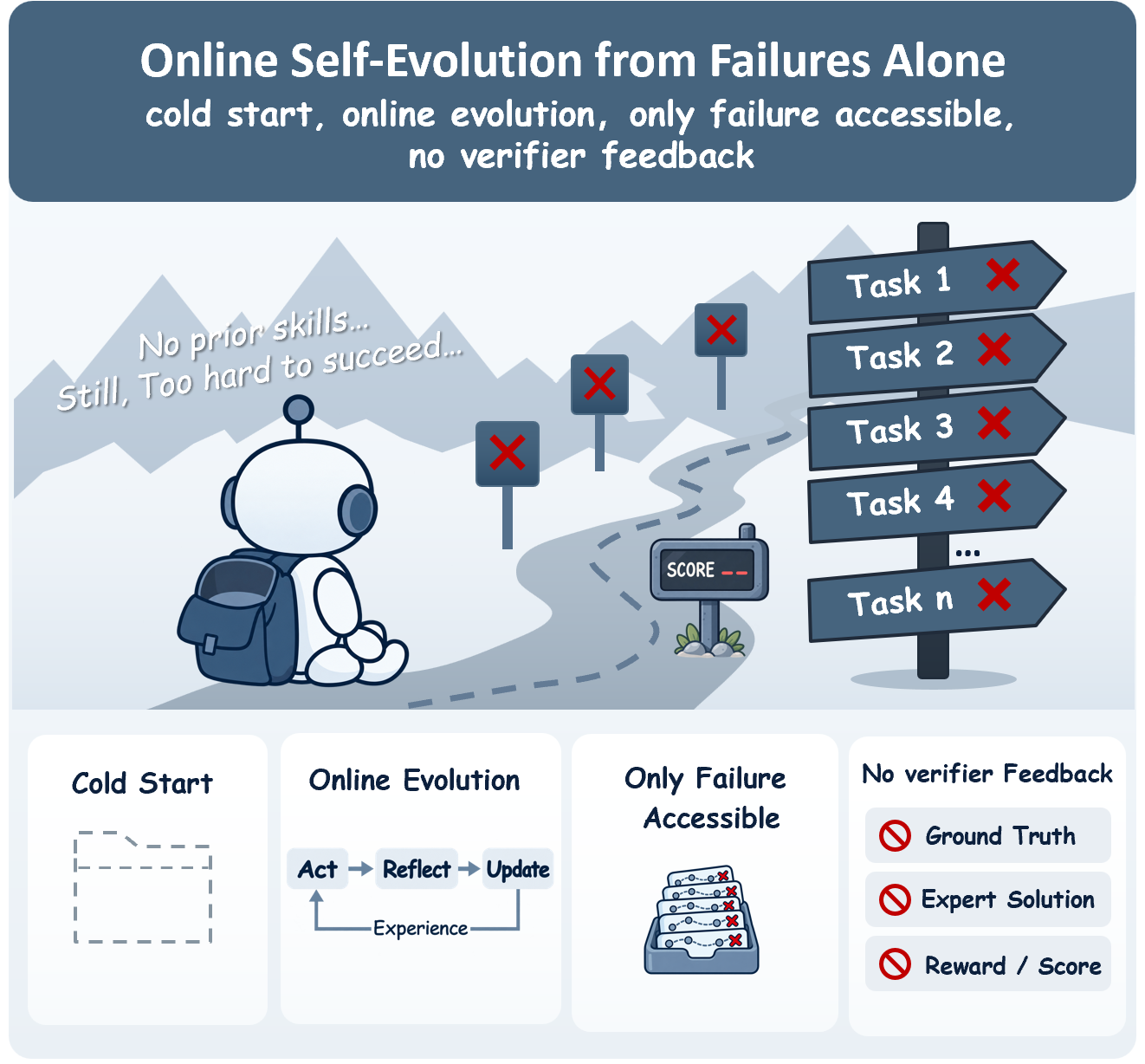}
  \caption{\emph{Cold-Start Failure-Only Evolution} on a backbone-specific, failure-dense online task stream. Starting from an empty external state, agents learn only from failed interactions, without access to ground-truth answers, expert solutions, or verifier feedback. We investigate whether existing self-evolving agents can turn failure-derived experience into transferable capability under these conditions.}
  \label{fig:failure_only_overview}
\end{figure}

Yet existing self-evolution studies typically assume access to positive learning signals, such as successful trajectories, verifier feedback, or already useful experience \citep{ouyang2026reasoningbank,yang2026skillforge,skillopt2026}. In challenging open environments, however, these signals may be scarce precisely when adaptation is most needed: agents can start without prior experience, repeatedly fail on unfamiliar tasks, and receive little guidance on how to improve. This regime provides a critical test of autonomous self-evolution---whether agents can bootstrap new capabilities from failure itself, rather than relying on success as a prerequisite for learning.

To study this problem, we introduce \emph{Cold-Start Failure-Only Evolution} (CFOE), a setting that isolates self-evolution from failures themselves (Figure~\ref{fig:failure_only_overview}). Agents start from an empty external state and evolve over a failure-dense online task stream. Successful executions may occur, but they are used only for evaluation and never for reflection or updating. Verifier feedback, reward scores, ground-truth answers, and expert solutions are also hidden from the agent and updater. Thus, self-evolution is driven exclusively by failed interactions, requiring agents to diagnose failures and extract reusable experience without privileged guidance.

We first investigate \textbf{(RQ1)} \textit{to what extent agents can learn from failures under Cold-Start Failure-Only Evolution, and whether such learning generalizes to unseen tasks} (\S\ref{sec:main_experiments}). We evaluate three representative paradigms---ReasoningBank for reasoning-memory evolution \citep{ouyang2026reasoningbank}, SkillOpt for skill evolution \citep{skillopt2026}, and Evo-Harness for runtime harness evolution \citep{evoharness2026}---across multiple LLM families and four interactive environments. Across frameworks, models, and environments, we observe a consistent \emph{repair--generalization gap}: failure-derived experience can partially recover previously failed tasks, yet these gains rarely translate into sustained improvements on unseen tasks and may even degrade held-out performance. Extending evolution to much longer horizons does not close this gap.

We then investigate \textbf{(RQ2)} \textit{what prevents failure-derived experience from becoming transferable capability} (\S\ref{sec:analysis}). By tracing trajectory interpretation, failure attribution, and experience consolidation, we identify recurring errors such as false-success attribution, misattribution, distorted consolidation, and harmful guidance. These results reveal a diagnosis bottleneck: agents may turn plausible but unsupported interpretations of failures into persistent experience. Yet diagnosis alone is not the whole problem. Even after retaining only verified-correct experience, source-task recovery improves while held-out performance remains unchanged, showing that correct experience can still remain overly tied to the source trajectory and fail to transfer.

Finally, we turn to \textbf{(RQ3)} \textit{what learning conditions are missing for effective experience transfer} (\S\ref{sec:learning_conditions}). We relax CFOE by allowing successful trajectories, exposing verifier feedback, or permitting additional interaction during reflection. Verifier feedback and extra interaction both improve failure understanding, but yield limited transfer gains. In contrast, successful interactions produce substantially stronger held-out improvements, suggesting that positive behavioral evidence provides information that failure analysis alone cannot easily recover. Failures reveal what went wrong, while successful trajectories show what reliably works and how useful actions compose into complete procedures.

Overall, our study provides a controlled evaluation of self-evolution before success. We show that current agents can extract useful local corrections from failed interactions, but struggle to turn them into transferable capability (\S\ref{sec:main_experiments}). We further trace this limitation to unreliable failure interpretation and the limited transferability of even correct experience (\S\ref{sec:analysis}), and show that successful behavioral evidence provides substantially stronger support for transferable evolution than richer failure diagnosis alone (\S\ref{sec:learning_conditions}). These findings suggest that current self-evolving agents are better at \emph{repairing observed failures} than at \emph{building generalizable capabilities from failure}, highlighting a fundamental challenge for autonomous learning in open environments.

\section{Cold-Start Failure-Only Evolution}

\wx{
We formulate an online self-evolving agent by a fixed backbone policy $\pi_\theta$, an evolving external state $S_t$, an updater $U$, and a verifier $V$. Given a sequential task stream $\{x_t\}_{t=1}^{T}$, the agent executes each task with its current external state,
}
\[
\tau_t \sim \pi_\theta(\cdot \mid x_t,S_{t-1}),
\]
\wx{
where $\theta$ remains fixed throughout evolution. The updater then modifies $S_t$ based on the interaction trajectory. Existing self-evolving agents mainly differ in the representation of $S_t$---e.g., reasoning memory, agent skills, or runtime harness state---and in how $U$ updates it.
}

\wx{
Our \emph{Cold-Start Failure-Only Evolution} (CFOE) setting imposes three constraints. Evolution starts from an empty external state, only failed trajectories are used for updating, and all verifier feedback is hidden from both the agent and the updater. Let $y_t=V(x_t,\tau_t)\in\{0,1\}$ denote the verifier judgment available only to the experimental controller. Then
}
\[
S_t =
\begin{cases}
S_{t-1}, & y_t=1,\\
U(S_{t-1},\tau_t), & y_t=0.
\end{cases}
\]
\wx{
Thus, successful executions may occur occasionally but never contribute to evolution, while success/failure labels, reward scores, verifier explanations, ground-truth answers, and expert solutions remain inaccessible throughout the process.
}

\section{\wx{Failure-Only Evolution Repairs but Does Not Generalize}}
\label{sec:main_experiments}

\wx{We begin by empirically answering \textbf{RQ1}: to what extent can agents learn from failures, and whether such learning generalizes to unseen tasks.}

\subsection{Experimental Setup}
\label{subsec:experimental_setup}

\paragraph{Self-Evolving Agents}
We study three paradigms with frozen backbones: reasoning-memory evolution with \textbf{ReasoningBank} \citep{ouyang2026reasoningbank}, skill evolution with \textbf{SkillOpt} \citep{skillopt2026}, and runtime harness evolution with \textbf{Evo-Harness} \citep{evoharness2026}. In all three frameworks, self-evolution arises from the accumulation and refinement of external artifacts. Framework details are provided in Appendix~\ref{app:agents}.

\paragraph{Backbone Model.}
We conducted experiments with 10 LLM backbones that span multiple model families, parameter scales, and architectures. Across all three self-evolution frameworks, we include a wide spectrum of dense variants of Qwen3.5 family (4B, 9B, and 27B) \citep{qwen3.5} and Gemma 4 family (Gemma-4-12B-it and Gemma-4-31B-it) \citep{gemmateam2026gemma4}, as well as the mixture-of-experts models Qwen3.5-35B-A3B and Gemma-4-26B-A4B-it. We further extend the evaluation to DeepSeek-V4-Pro \citep{deepseekai2026deepseekv4}, GLM-5.1 \citep{glm5team2026glm5vibecodingagentic} and GPT-5.6 Sol \citep{openai2026gpt56} in selected frameworks, given the cost of the held-out evaluation at multiple evolution checkpoints.

\paragraph{Environment \& Benchmark.}
Our experiments cover household embodied task execution, long-horizon interactive tool use, online shopping, and web navigation across four \emph{quality-verified} environments: \textbf{ALFWorld} \citep{shridhar2021alfworld}, \textbf{\texorpdfstring{$\bm{\tau}^2$}{tau2}-Bench-Verified} \citep{barres2026tau2,cuadron2025saber}, \textbf{WebArena-Verified} \citep{hattami2025webarena} (Shopping, Shopping Admin, Reddit, and GitLab), and \textbf{WebShop} \citep{yao2022webshop}. These environments require multi-step interaction and provide a diverse testbed for long-horizon agentic adaptation. Benchmark details can be found in the Appendix \ref{app:benchmarks}.

\paragraph{Implementation Details.}
Closed-source and large-scale open-weight backbones are accessed via official APIs, while other open-weight models are deployed locally with vLLM \citep{kwon2023efficient} on NVIDIA H100 GPUs. We adopt the recommended configurations and evaluation protocols for each agent framework and benchmark. For SkillOpt, we adapt the offline setup to an online setting to align with our self-evolution protocol. Further configurations are provided in the Appendix~\ref{app:implementation-execution-details}.

\paragraph{Failure-Dense Online Evolution Streams Construction.}
Given the inherent differences in initial task-solving capabilities across model backbones, we first construct a challenging task stream for each backbone on each benchmark independently for online self-evolution. Each task involved in the constructed stream has resulted in a consistent failure in 3 independent preliminary rollouts with the corresponding backbone, inducing a high density of failures during subsequent online self-evolution. Details of the stream construction procedure are provided in the Appendix~\ref{app:failure_dense_stream_construction}.

\paragraph{Online Evolution from Failures Alone}
Each run starts with no accumulated memory or skills and a fresh harness state. The agent processes the challenging task stream sequentially, using failed trajectories to reflect and evolve its memory, agent skills, or runtime harness state. Occasional successes are recorded only for final reporting; the agent moves directly to the next task without using them for reflection or self-evolution updates.

\paragraph{Ground Truth \& Reward Blind}
The controller uses verifier judgments to collect failed trajectories, but neither the agent nor updater receives outcome labels, rewards, verifier explanations, ground-truth answers, or expert solutions. Agents are also unaware of this failure-only selection. Details of feedback setting appear in Appendix~\ref{app:verifier_feedback_blind}.

\begin{figure*}[t]
  \centering
  \includegraphics[width=0.90\textwidth]{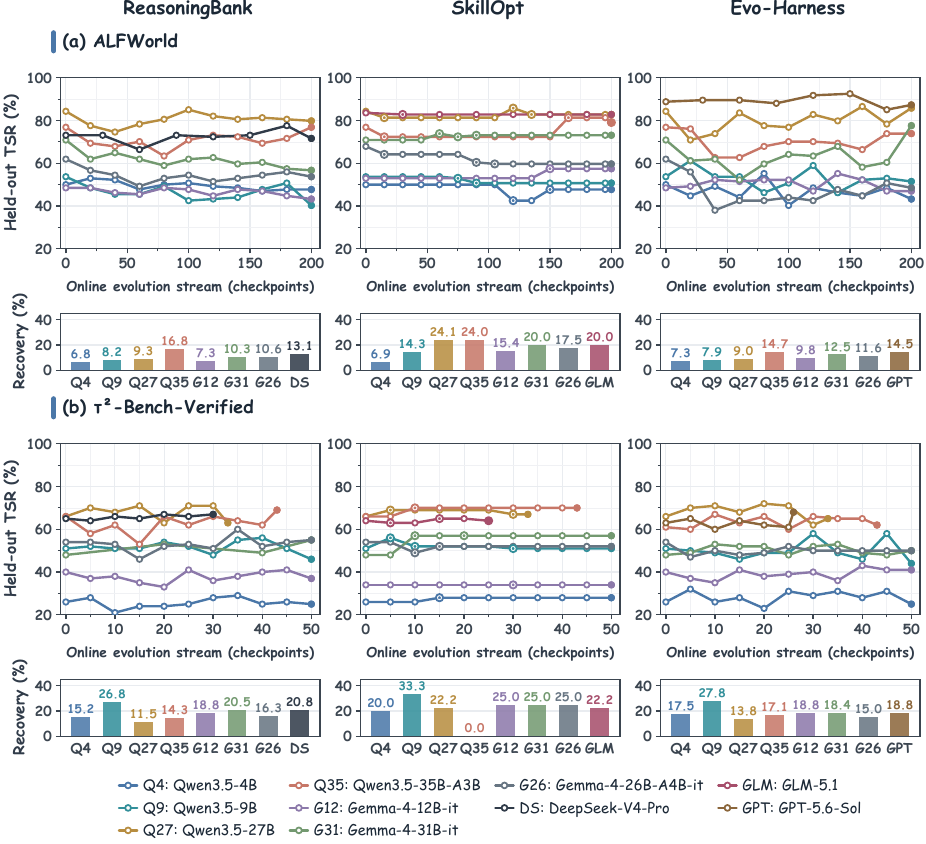}
  \caption{Held-out task success rate (TSR) and source-task recovery during cold-start failure-only evolution on (a) ALFWorld and (b) $\tau^2$-Bench-Verified. Results are shown for ReasoningBank, SkillOpt, and Evo-Harness, with eight backbone models per framework. Some $\tau^2$-Bench-Verified failure-dense streams contain fewer than 50 tasks.}
  \label{fig:failure_only_main}
\end{figure*}

\begin{figure*}[t]
  \centering
  \includegraphics[width=0.93\textwidth]{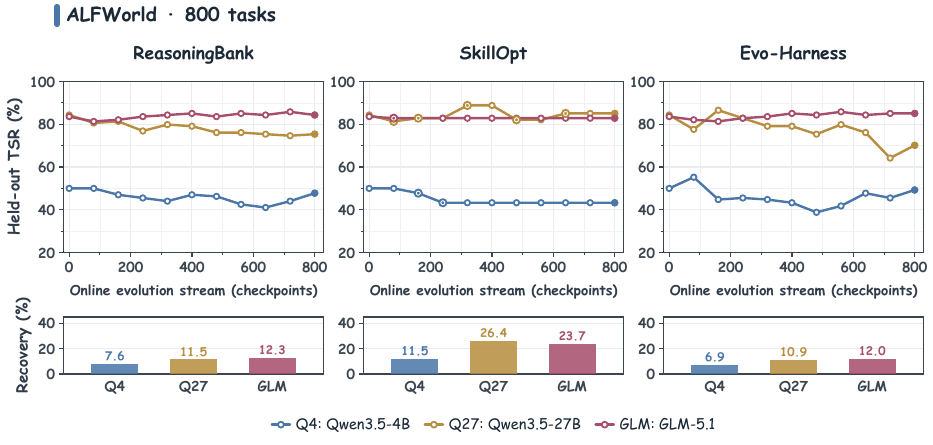}
  \caption{Held-out TSR and source-task recovery during cold-start failure-only evolution over 800 ALFWorld tasks. Results are shown for Qwen3.5-4B, Qwen3.5-27B, and GLM-5.1 across ReasoningBank, SkillOpt, and Evo-Harness.}
  \label{fig:failure_only_long_horizon}
\end{figure*}

\subsection{\wx{Results and Analysis}}
\label{subsec:failure_only_evaluation}

Figure~\ref{fig:failure_only_main} presents CFOE on ALFWorld and $\tau^2$-Bench-Verified across the three frameworks. Each benchmark pairs held-out curves with source-task recovery bars for the same backbones. Results on WebArena-Verified and WebShop are provided in Appendix~\ref{app:additional_main_results}.

\paragraph{Cold-Start Failure-Only Evolution yields limited held-out gains and can even degrade performance.}
We evaluate successive checkpoints on a fixed unseen held-out set disjoint from the online stream. Agent artifacts remain unchanged throughout evaluation, with no further self-evolution on held-out tasks (Appendix~\ref{app:held_out_evaluation}). The held-out curves in Figure~\ref{fig:failure_only_main} show limited held-out gains, with some configurations even declining as experience accumulates. SkillOpt exhibits smaller held-out fluctuations than ReasoningBank and Evo-Harness. This relative stability may stem from its validation-based acceptance mechanism, which retains candidate skills only when they improve validation performance. Even so, gains on held-out tasks remain limited.

\paragraph{Failure-derived experience supports partial recovery of original failed tasks.}
After each effective update, we retry its failed source tasks once from their initial states, using the experience state immediately after that update without further reflection or updates. For these two benchmarks, recovery is the percentage of successful retries over the entire online run (Appendix~\ref{app:source_recovery_protocol}). The recovery bars in Figure~\ref{fig:failure_only_main} show partial source-task recovery, with rates varying across backbones, frameworks, and environments. Qwen3.5-27B with SkillOpt recovers more ALFWorld source tasks than Qwen3.5-4B, yet its held-out performance still declines. These results show that improvements in source-task recovery do not reliably carry over to held-out tasks.

\subsection{Long-Horizon Online Self-Evolution}
\label{subsec:long_horizon}

To examine whether these limitations persist as online self-evolution scales to a longer horizon, we extend the ALFWorld task stream to 800 episodes. All 800 tasks have failed in three independent preliminary rollouts with the corresponding backbone and no accumulated experience. Figure~\ref{fig:failure_only_long_horizon} pairs held-out curves and recovery performance for Qwen3.5-4B, Qwen3.5-27B, and GLM-5.1 across all three frameworks.

The limited held-out gains persist over the scaled evolution horizon. Held-out performance declines further for Qwen3.5-27B with Evo-Harness and plateaus for Qwen3.5-4B with SkillOpt, despite partial source-task recovery in both configurations. These long-horizon results are consistent with our earlier observations: failure-derived experience supports partial source-task recovery, but continued accumulation does not yield sustained held-out improvements.

\section{Causes of Limited Experience Transfer}
\label{sec:analysis}

\wx{
To answer \textbf{RQ2}---what prevents agents from turning failure-derived experience into transferable capability---we conduct two complementary analyses. We first examine the \emph{faithfulness of experience formation}, tracing how failed trajectories are interpreted, attributed, and consolidated into reusable experience (\S\ref{subsec:experience_correctness}). We then ask whether, even when the extracted experience is correct, agents can effectively utilize and transfer it to new tasks (\S\ref{subsec:correct_experience_transfer}).
}

\begin{table}[!t]
\centering
\small
\setlength{\tabcolsep}{2pt}
\begin{tabular*}{\linewidth}{@{\extracolsep{\fill}}llcccc@{}}
\toprule
\textbf{Model} & \textbf{Paradigm} & \makebox[2em]{\textbf{M1}} & \makebox[2em]{\textbf{M2}} & \makebox[2em]{\textbf{M3}} & \makebox[2em]{\textbf{M4}} \\
\midrule
\rowcolor{gray!8}
\multicolumn{6}{c}{\textit{ALFWorld}} \\
\midrule
\texttt{Qwen3.5-27B} & \texttt{Memory} & 41.3 & 46.5 & 32.6 & 35.5 \\
\texttt{Qwen3.5-27B} & \texttt{Skill} & 43.8 & 39.7 & 28.8 & 37.0 \\
\texttt{Qwen3.5-27B} & \texttt{Harness} & 29.9 & 38.6 & 31.5 & 36.2 \\
\texttt{GPT-5.6-Sol} & \texttt{Harness} & 21.2 & 31.0 & 25.7 & 23.9 \\
\midrule
\rowcolor{gray!8}
\multicolumn{6}{c}{\textit{$\tau^2$-Bench-Verified}} \\
\midrule
\texttt{Qwen3.5-27B} & \texttt{Memory} & 34.0 & 53.2 & 29.8 & 42.6 \\
\texttt{Qwen3.5-27B} & \texttt{Skill} & 37.8 & 48.6 & 35.1 & 43.2 \\
\texttt{Qwen3.5-27B} & \texttt{Harness} & 27.9 & 44.2 & 34.9 & 32.6 \\
\texttt{GPT-5.6-Sol} & \texttt{Harness} & 31.7 & 39.0 & 19.5 & 26.8 \\
\midrule
\rowcolor{gray!8}
\multicolumn{6}{c}{\textit{WebShop}} \\
\midrule
\texttt{Qwen3.5-27B} & \texttt{Skill} & 43.3 & 31.3 & 23.9 & 38.8 \\
\texttt{Qwen3.5-27B} & \texttt{Harness} & 28.2 & 32.4 & 23.9 & 36.6 \\
\texttt{GPT-5.6-Sol} & \texttt{Harness} & 19.7 & 24.6 & 26.2 & 23.0 \\
\bottomrule
\end{tabular*}
\caption{Percentage of reviewed updates exhibiting each failure mode. A single update may exhibit multiple failure modes.}
\label{tab:experience_formation_errors}
\end{table}

\subsection{Errors in Experience Formation}
\label{subsec:experience_correctness}

To identify the causes of limited experience transfer, we manually inspect experience accumulated during self-evolution. For each evolution update, we trace how the agent interprets the failed trajectory, how this interpretation informs failure attribution, and how the resulting explanation is consolidated into experience. We analyze Qwen3.5-27B with ReasoningBank (Memory), SkillOpt (Skill), and Evo-Harness (Harness), and GPT-5.6-Sol with Evo-Harness. ReasoningBank is evaluated on ALFWorld and $\tau^2$-Bench-Verified; SkillOpt and Evo-Harness are evaluated on ALFWorld, $\tau^2$-Bench-Verified, and WebShop.

We categorize reasons for incorrect experience formation into four types:
\textbf{(M1) False-Positive Success Attribution} (False Success), where actions or intermediate states that contributed to failure are credited as successful practices (e.g., treating a budget-wasting exhaustive search as a successful search strategy).
\textbf{(M2) Misidentified \& Hallucinated Failure Attribution} (Misattribution), where relevant causes are overlooked or replaced with unrelated factors or unsupported explanations (e.g., falsely interpreting filters on the wrong category page as evidence that the requested category has been reached).
\textbf{(M3) Distorted Experience Consolidation} (Distortion), where observations or inferences lose their conditions or acquire unwarranted certainty when converted into experience (e.g., turning ``go back before a new search'' into ``always go back from a results page'').
\textbf{(M4) Poisonous Future Behavior Guidance} (Harmful Guidance), where incorrect or unsupported judgments become direct instructions that badly conflict with environmental constraints and can lead to further failures (e.g., advising agents to abandon troubleshooting and escalate after three unsuccessful explanations, even when repairs remain available).
Some instances may exhibit more than one failure mode. Detailed annotation criteria and cases are provided in Appendix~\ref{app:annotation}.

Table~\ref{tab:experience_formation_errors} summarizes the distribution of these errors across models and benchmarks. Misattribution is particularly common in $\tau^2$-Bench Verified, whereas False Success is more prominent in ALFWorld. Stronger models generally exhibit fewer errors, and harness evolution shows lower error rates than memory or skill evolution. These error patterns reveal a weakness in learning under \emph{Cold-Start Failure-Only Evolution}. Unsupported interpretations of failed attempts may be consolidated into reusable experience and carried forward as action rules, potentially guiding subsequent behavior in ways that conflict with task constraints.

\begin{figure*}[t]
  \centering
  \includegraphics[width=0.95\textwidth]{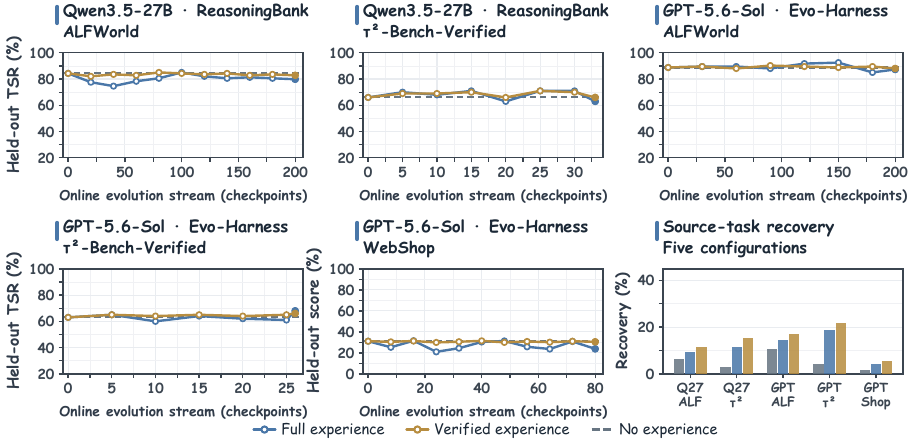}
  \caption{Held-out performance and source-task recovery with full experience, verified experience, and no experience. Dashed lines denote the no-experience held-out baseline.}
  \label{fig:correct_experience_transfer}
\end{figure*}

\subsection{Utilization and Transfer of Correct Experience}
\label{subsec:correct_experience_transfer}

\paragraph{Setup.} We further investigate whether agents can effectively use failure-derived experience to improve performance on new tasks, even when the experience itself is correct. To this end, we retain only manually verified memory entries or harness modifications at each checkpoint. We evaluate Qwen3.5-27B with ReasoningBank on ALFWorld and $\tau^2$-Bench-Verified, and GPT-5.6-Sol with Evo-Harness on these benchmarks and WebShop. Full experience, verified experience, and no experience are compared on the same source and held-out tasks, with experience fixed during evaluation.

\paragraph{Results \& Analysis.} Even when the experience extracted from failures is correct, agents still struggle to apply it effectively to held-out tasks. As shown in Figure~\ref{fig:correct_experience_transfer}, retaining only verified experience modestly improves source-task recovery in all five configurations, yet final held-out scores remain at or below the no-experience baseline in four.

These results further indicate the challenge of learning transferable experience under \emph{Cold-Start Failure-Only Evolution}. Even when the experience is correct in terms of content, agents still struggle to use it effectively to form transferable capability. Our inspection shows that such experience remains tied to specific steps and conditions of the source task, indicating a gap in extracting generalizable knowledge from harsh and dense failures.

\section{Effects of Learning Conditions on Experience Transfer}
\label{sec:learning_conditions}

\wx{We next address \textbf{RQ3}: what learning conditions are missing for agents to acquire more transferable experience? We systematically relax the CFOE setting along three dimensions by introducing successful experience, exposing verifier feedback, and allowing additional environment interaction during failure reflection.}

\paragraph{Setup.}
We adopt the three self-evolving frameworks with Qwen3.5-27B and evaluate held-out performance on the environments described in \S\ref{subsec:experimental_setup}. 
\wx{
We first vary the source of trajectories available for experience extraction:
}
\textbf{(1) Fail}: learns only from failed interactions in the failure-dense task stream, following the original setting in \S\ref{sec:main_experiments}.
\textbf{(2) Success}: samples tasks that succeeded in all three preliminary rollouts and learns only from success during online evolution.
\textbf{(3) Mix}: randomly samples tasks and allows learning from both successful and failed interactions during online evolution.
Task streams are sampled separately for each setting. We evaluate Success and Mix only with ReasoningBank and SkillOpt, as Evo-Harness retains its failure-focused update procedure.
\wx{
In these comparisons, official verifier judgments remain hidden from the agent and updater, which we denote as \textbf{Blind}.
}

\wx{
We next isolate the effect of external evaluative signals by exposing the official verifier feedback under failure-only evolution (\textbf{Feedback}), including outcome labels, reward scores, and task-specific feedback available in each environment.
}

\wx{
Finally, we examine whether additional interaction can compensate for missing information in a failed trajectory. Inspired by the agentic error analysis in Trace2Skill~\citep{ni2026trace2skilldistilltrajectorylocallessons}, we allow the agent to resume interaction with the environment during reflection (\textbf{Resume}) to diagnose the failure and explore possible corrections. Reflection is still triggered by the original failed rollout, but the agent may use observations collected during these additional interactions when extracting experience, including those from a subsequently successful attempt. Verifier judgments and ground truth remain hidden. We allocate 10 interaction turns followed by one tool-free summary turn in the main comparison. Additional settings and budget comparisons are provided in Appendix~\ref{app:ablations}.
}

\begin{figure*}[t]
  \centering
  \includegraphics[width=0.9\textwidth]{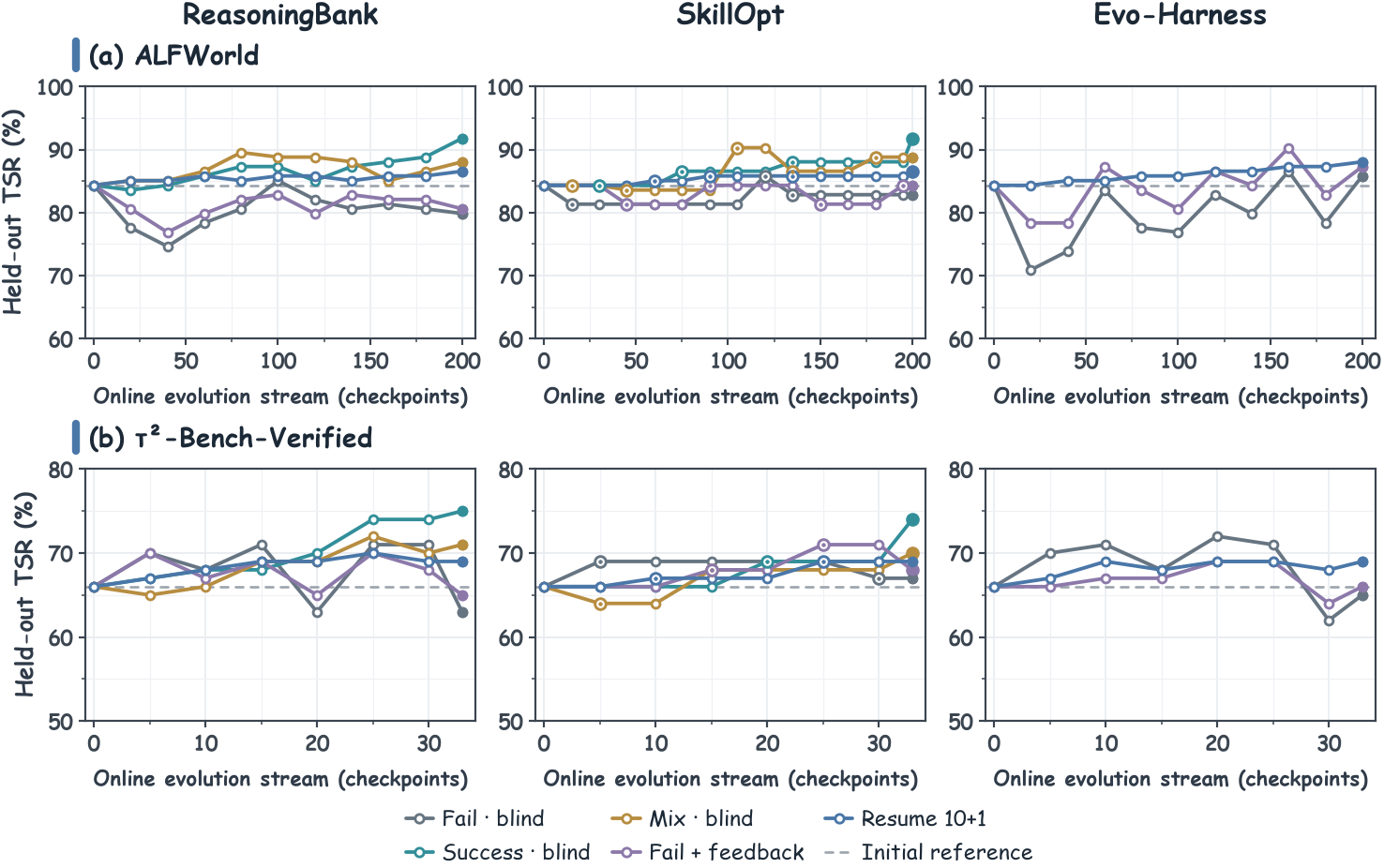}
  \caption{Learning-condition comparisons on ALFWorld and $\tau^2$-Bench-Verified with Qwen3.5-27B. Fail + Blind reproduces the reported failure-only baseline; Resume uses 10 interaction turns and one summary turn. Dashed lines indicate initial held-out performance.}
  \label{fig:learning_conditions_main}
\end{figure*}

\paragraph{Results \& Analysis.}
Figure~\ref{fig:learning_conditions_main} reports held-out performance during online self-evolution under different learning conditions on ALFWorld and $\tau^2$-Bench-Verified. Appendix~\ref{app:ablations} provides additional results on WebArena-Verified and WebShop. 
\wx{
It further compares 5, 10, and 20 additional interaction turns for Resume, and evaluates verifier feedback under successful and mixed trajectory streams.
}

\paragraph{Successful interactions support more effective experience transfer.}
As shown in Figure~\ref{fig:learning_conditions_main}, learning from successful interactions yields larger held-out improvements than learning exclusively from failures. Mix also benefits from incorporating successful interactions, although its gains are generally smaller than those of Success. Successful trajectories provide examples of action sequences that complete a task, giving agents concrete references when extracting reusable experience. These results suggest that access to successful executions is important for learning transferable experience.

\paragraph{Verifier feedback improves failure-only learning, but its benefits remain limited.}
Providing official verifier judgments improves held-out performance relative to Fail + Blind. The gains nevertheless remain limited compared with those from successful interactions. Feedback supplies additional evidence about task outcomes and unmet requirements, but agents must still infer behavioral corrections and express them as reusable experience. The remaining transfer difficulties suggest that external judgment only partially addresses the challenges of learning from failed interactions.

\paragraph{Interactive failure diagnosis offers further learning opportunities with limited transfer gains.}
Revisiting failed tasks during reflection also improves held-out performance over the CFOE baseline. Through additional interactions, agents can investigate uncertain steps and explore possible corrections before consolidating experience. These opportunities yield some transferable gains, although the overall improvement remains limited. Together with the feedback results, this finding suggests that additional failure evidence can help learning, while extracting transferable experience remains challenging.

\section{Related Work}
\label{sec:related_work}
\paragraph{Experience-Driven Self-Evolving Agents.}
Recent work explores agents that improve with frozen model parameters by retaining and reusing interaction experience \citep{wang2023voyager,zhang2026ace}. Experience can be collected offline or accumulated during deployment. Offline, AutoGuide extracts context-aware guidelines from prior trajectories \citep{fu2024autoguide}. Online approaches update environmental rules through interaction \citep{chen2024automanual} or maintain evolving memories of strategies and code for subsequent queries \citep{suzgun2025dynamiccheatsheet}. Voyager stores executable skills \citep{wang2023voyager}, while Agent Workflow Memory induces reusable workflows \citep{wang2025awm}. Recent frameworks distill reasoning memories \citep{ouyang2026reasoningbank}, refine skill instructions \citep{skillopt2026}, and compile harness guidance \citep{evoharness2026}. ACE organizes accumulated strategies into structured playbooks through incremental context updates \citep{zhang2026ace}. We examine memory, skill, and harness evolution from a shared cold start as agents learn from failures on initially challenging tasks.

\paragraph{Learning from Failure.}
Failure learning has been studied through reflection, feedback, and experience reuse. Reflexion stores reflections on task feedback to guide subsequent trials \citep{shinn2023reflexion}, while ExpeL contrasts failed and successful executions to derive cross-task rules \citep{zhao2024expel}. Output revision uses model-generated feedback in Self-Refine \citep{madaan2023selfrefine} and tool-based feedback in CRITIC \citep{gou2024critic}. For persistent experience updates, ReasoningBank builds memories from self-judged trajectories \citep{ouyang2026reasoningbank}, and ERL examines failure-derived heuristics and extraction without outcome feedback \citep{allard2026erl}. Other approaches check updates against additional evidence. HarnessFix grounds diagnoses in execution traces and harness artifacts and validates proposed repairs \citep{chen2026harnessfix}; Sentry retains recovery lessons after checking subsequent interactions and evaluates reuse on unseen tasks without task-level rewards \citep{ji2026sentry}. These studies connect failure diagnosis to subsequent behavior across feedback conditions. We study failure-dense learning from unsuccessful interactions without verifier explanations, separately evaluating source-task repair and held-out transfer.

\section{Conclusion}

We study whether self-evolving LLM agents can acquire transferable capability from failure alone through \emph{Cold-Start Failure-Only Evolution} (CFOE). Across multiple self-evolving paradigms, LLM backbones, and interactive environments, we uncover a consistent \emph{repair--generalization gap}: failure-derived experience can repair observed failures but rarely transfers to unseen tasks. Further analyses show that both unreliable failure interpretation and limited transfer of even correct experience contribute to this gap. Successful trajectories provide substantially stronger transfer gains than verifier feedback or additional failure exploration, suggesting that current agents are better at learning \emph{what went wrong} than \emph{what reliably works}.

\section*{Limitations}

While our study provides a systematic investigation into the limitations of cold-start failure-only evolution in LLM agents, several limitations remain. First, our evaluation prioritizes challenging tasks from a focused set of benchmarks that span both web-based and embodied scenarios. However, these benchmarks may not fully capture the diversity of difficult situations agents encounter in real-world deployment environments, especially those involving multi-agent interactions or multi-modal inputs. Extending our analysis to broader distributions of challenging tasks remains an important direction. Second, our experiments study only scale self-evolving agents over a finite number of self-evolution steps, with the longest evaluation covering 800 tasks on ALFWorld. While this already reveals persistent limitations in experience transfer, real-world deployed agents may undergo self-evolution over far longer, and potentially unbounded, time horizons. How experience transfer evolves under such indefinite failure accumulation, and whether transferable gains emerge beyond the studied regime, remain open questions for future work and more scaling exploration.

Overall, this work provides an empirical investigation into the limits of learning from failures in self-evolving agents. We hope future efforts will explore more general, principled, and verifiable mechanisms to support transferable learning from failures in experience-driven AI systems.

\section*{Ethical Considerations}

This work is conducted solely for research purposes, with the goal of understanding the limitations of learning from failures in self-evolving agents. All experiments are performed in controlled benchmark environments, with restricted learning signals used to study experience formation and transfer. We believe that systematically identifying and characterizing errors in accumulated experience is important for reducing the risk of repeated mistakes that could affect users in real-world applications. By exposing failure modes and transfer limitations in current self-evolving frameworks, this work aims to inform the design of evaluation and oversight mechanisms for more reliable and responsible agentic systems.

\bibliography{custom}

@inproceedings{shridhar2021alfworld,
  title = {{ALFWorld}: Aligning Text and Embodied Environments for Interactive Learning},
  author = {Shridhar, Mohit and Yuan, Xingdi and C{\^o}t{\'e}, Marc-Alexandre and Bisk, Yonatan and Trischler, Adam and Hausknecht, Matthew},
  booktitle = {International Conference on Learning Representations},
  year = {2021},
  url = {https://arxiv.org/abs/2010.03768},
}

@inproceedings{cuadron2025saber,
  title = {{SABER}: Small Actions, Big Errors---Safeguarding Mutating Steps in {LLM} Agents},
  author = {Cuadron, Alejandro and Yu, Pengfei and Liu, Yang and Gupta, Arpit},
  booktitle = {ICLR 2026 Workshop on Memory for LLM-Based Agentic Systems (MemAgents)},
  year = {2026},
  url = {https://openreview.net/forum?id=En2z9dckgP},
}

@inproceedings{hattami2025webarena,
  title = {{WebArena Verified}: Reliable Evaluation for Web Agents},
  author = {El Hattami, Amine and Thakkar, Megh and Chapados, Nicolas and Pal, Christopher},
  booktitle = {Workshop on Scaling Environments for Agents},
  year = {2025},
  url = {https://openreview.net/forum?id=94tlGxmqkN},
}

@inproceedings{yao2022webshop,
  author = {Yao, Shunyu and Chen, Howard and Yang, John and Narasimhan, Karthik},
  booktitle = {Advances in Neural Information Processing Systems},
  doi = {10.52202/068431-1508},
  editor = {S. Koyejo and S. Mohamed and A. Agarwal and D. Belgrave and K. Cho and A. Oh},
  pages = {20744--20757},
  publisher = {Curran Associates, Inc.},
  title = {{WebShop}: Towards Scalable Real-World Web Interaction with Grounded Language Agents},
  url = {https://proceedings.neurips.cc/paper_files/paper/2022/file/82ad13ec01f9fe44c01cb91814fd7b8c-Paper-Conference.pdf},
  volume = {35},
  year = {2022},
}

@inproceedings{ouyang2026reasoningbank,
  title = {{ReasoningBank}: Scaling Agent Self-Evolving with Reasoning Memory},
  author = {Siru Ouyang and Jun Yan and I-Hung Hsu and Yanfei Chen and Ke Jiang and Zifeng Wang and Rujun Han and Long T. Le and Samira Daruki and Xiangru Tang and Vishy Tirumalashetty and George Lee and Mahsan Rofouei and Hangfei Lin and Jiawei Han and Chen-Yu Lee and Tomas Pfister},
  booktitle = {International Conference on Learning Representations},
  year = {2026},
  url = {https://proceedings.iclr.cc/paper_files/paper/2026/hash/980ea04d23d1f6908964eba2a74afe45-Abstract-Conference.html},
  volume = {2026},
  pages = {94327--94354},
}

@article{skillopt2026,
  title = {{SkillOpt}: Executive Strategy for Self-Evolving Agent Skills},
  author = {Yang, Yifan and Gong, Ziyang and Huang, Weiquan and Yang, Qihao and Zhou, Ziwei and Huang, Zisu and Li, Yan and Gao, Xuemei and Dai, Qi and Liu, Bei and Qiu, Kai and Yang, Yuqing and Chen, Dongdong and Yang, Xue and Luo, Chong},
  journal = {arXiv preprint arXiv:2605.23904},
  year = {2026},
  url = {https://arxiv.org/abs/2605.23904},
  doi = {10.48550/arXiv.2605.23904},
}

@article{evoharness2026,
  title = {{Evo-Harness}: Context-to-Harness Skill Compilation for Self-Evolving Agents},
  author = {Wei, Tianxin and Shi, Zhan and Lin, Minhua and He, Bing and Liu, Zewen and Sang, Yisi and Bei, Yuanchen and Ning, Xuying and Zou, Jiaru and Li, Ting-Wei and Lin, Xiao and Zhao, Yanjun and Wang, Chi and Dumoulin, Benoit and Wang, Dakuo and He, Jingrui and Lu, Hanqing},
  journal = {arXiv preprint arXiv:2608.15071},
  year = {2026},
  url = {https://arxiv.org/abs/2608.15071},
  doi = {10.48550/arXiv.2608.15071},
  note = {Accepted to {EMNLP} 2026 (main conference)},
}

@inproceedings{wang2025awm,
  title = {Agent Workflow Memory},
  author = {Wang, Zora Zhiruo and Mao, Jiayuan and Fried, Daniel and Neubig, Graham},
  booktitle = {Proceedings of the 42nd International Conference on Machine Learning},
  pages = {63897--63911},
  year = {2025},
  volume = {267},
  series = {Proceedings of Machine Learning Research},
  publisher = {PMLR},
  url = {https://proceedings.mlr.press/v267/wang25bx.html},
}

@inproceedings{suzgun2025dynamiccheatsheet,
  title = {Dynamic Cheatsheet: Test-Time Learning with Adaptive Memory},
  author = {Suzgun, Mirac and Yuksekgonul, Mert and Bianchi, Federico and Jurafsky, Dan and Zou, James},
  booktitle = {Proceedings of the 19th Conference of the European Chapter of the Association for Computational Linguistics (Volume 1: Long Papers)},
  year = {2026},
  url = {https://aclanthology.org/2026.eacl-long.333/},
  pages = {7080--7106},
  publisher = {Association for Computational Linguistics},
  doi = {10.18653/v1/2026.eacl-long.333},
}

@inproceedings{madaan2023selfrefine,
  author = {Madaan, Aman and Tandon, Niket and Gupta, Prakhar and Hallinan, Skyler and Gao, Luyu and Wiegreffe, Sarah and Alon, Uri and Dziri, Nouha and Prabhumoye, Shrimai and Yang, Yiming and Gupta, Shashank and Majumder, Bodhisattwa Prasad and Hermann, Katherine and Welleck, Sean and Yazdanbakhsh, Amir and Clark, Peter},
  booktitle = {Advances in Neural Information Processing Systems},
  doi = {10.52202/075280-2019},
  pages = {46534--46594},
  publisher = {Curran Associates, Inc.},
  title = {{Self-Refine}: Iterative Refinement with Self-Feedback},
  url = {https://proceedings.neurips.cc/paper_files/paper/2023/file/91edff07232fb1b55a505a9e9f6c0ff3-Paper-Conference.pdf},
  volume = {36},
  year = {2023},
}

@inproceedings{gou2024critic,
  title = {{CRITIC}: Large Language Models Can Self-Correct with Tool-Interactive Critiquing},
  author = {Gou, Zhibin and Shao, Zhihong and Gong, Yeyun and Shen, Yelong and Yang, Yujiu and Duan, Nan and Chen, Weizhu},
  booktitle = {International Conference on Learning Representations},
  year = {2024},
  url = {https://proceedings.iclr.cc/paper_files/paper/2024/hash/fef126561bbf9d4467dbb8d27334b8fe-Abstract-Conference.html},
  volume = {2024},
  pages = {57734--57811},
}

@inproceedings{shinn2023reflexion,
  author = {Shinn, Noah and Cassano, Federico and Gopinath, Ashwin and Narasimhan, Karthik and Yao, Shunyu},
  booktitle = {Advances in Neural Information Processing Systems},
  doi = {10.52202/075280-0377},
  pages = {8634--8652},
  publisher = {Curran Associates, Inc.},
  title = {{Reflexion}: language agents with verbal reinforcement learning},
  url = {https://proceedings.neurips.cc/paper_files/paper/2023/file/1b44b878bb782e6954cd888628510e90-Paper-Conference.pdf},
  volume = {36},
  year = {2023},
}

@inproceedings{fu2024autoguide,
  author = {Fu, Yao and Kim, Dong-Ki and Kim, Jaekyeom and Sohn, Sungryull and Logeswaran, Lajanugen and Bae, Kyunghoon and Lee, Honglak},
  booktitle = {Advances in Neural Information Processing Systems},
  doi = {10.52202/079017-3811},
  pages = {119919--119948},
  publisher = {Curran Associates, Inc.},
  title = {{AutoGuide}: Automated Generation and Selection of Context-Aware Guidelines for Large Language Model Agents},
  url = {https://proceedings.neurips.cc/paper_files/paper/2024/file/d8efbb5dd415974eb095c3f06bff1f48-Paper-Conference.pdf},
  volume = {37},
  year = {2024},
}

@misc{ni2026trace2skilldistilltrajectorylocallessons,
  title = {{Trace2Skill}: Distill Trajectory-Local Lessons into Transferable Agent Skills},
  author = {Jingwei Ni and Yihao Liu and Xinpeng Liu and Yutao Sun and Mengyu Zhou and Pengyu Cheng and Dexin Wang and Erchao Zhao and Xiaoxi Jiang and Guanjun Jiang},
  year = {2026},
  eprint = {2603.25158},
  archiveprefix = {arXiv},
  primaryclass = {cs.AI},
  url = {https://arxiv.org/abs/2603.25158},
  doi = {10.48550/arXiv.2603.25158},
  note = {Accepted to {EMNLP} 2026 (main conference)},
}

@misc{qwen3.5,
  title = {{Qwen3.5}: Towards Native Multimodal Agents},
  author = {{Qwen Team}},
  month = {February},
  year = {2026},
  url = {https://qwen.ai/blog?id=qwen3.5},
}

@misc{gemmateam2026gemma4,
  title = {{Gemma 4} Technical Report},
  author = {{Gemma Team}},
  year = {2026},
  eprint = {2607.02770},
  archiveprefix = {arXiv},
  primaryclass = {cs.CL},
  url = {https://arxiv.org/abs/2607.02770},
  doi = {10.48550/arXiv.2607.02770},
}

@misc{glm5team2026glm5vibecodingagentic,
  title = {{GLM-5}: from Vibe Coding to Agentic Engineering},
  author = {{GLM-5-Team} and Aohan Zeng and Xin Lv and Zhenyu Hou and Zhengxiao Du and Qinkai Zheng and Bin Chen and Da Yin and Chendi Ge and Chenghua Huang and Chengxing Xie and Chenzheng Zhu and Congfeng Yin and Cunxiang Wang and Gengzheng Pan and Hao Zeng and Haoke Zhang and Haoran Wang and Huilong Chen and Jiajie Zhang and Jian Jiao and Jiaqi Guo and Jingsen Wang and Jingzhao Du and Jinzhu Wu and Kedong Wang and Lei Li and Lin Fan and Lucen Zhong and Mingdao Liu and Mingming Zhao and Pengfan Du and Qian Dong and Rui Lu and Li, Shuang and Shulin Cao and Song Liu and Ting Jiang and Xiaodong Chen and Xiaohan Zhang and Xuancheng Huang and Xuezhen Dong and Yabo Xu and Yao Wei and Yifan An and Yilin Niu and Yitong Zhu and Yuanhao Wen and Yukuo Cen and Yushi Bai and Zhongpei Qiao and Zihan Wang and Zikang Wang and Zilin Zhu and Ziqiang Liu and Zixuan Li and Bojie Wang and Bosi Wen and Can Huang and Changpeng Cai and Chao Yu and Chen Li and Chengwei Hu and Chenhui Zhang and Dan Zhang and Daoyan Lin and Dayong Yang and Di Wang and Ding Ai and Erle Zhu and Fangzhou Yi and Feiyu Chen and Guohong Wen and Hailong Sun and Haisha Zhao and Haiyi Hu and Hanchen Zhang and Hanrui Liu and Hanyu Zhang and Hao Peng and Hao Tai and Haobo Zhang and He Liu and Hongwei Wang and Hongxi Yan and Hongyu Ge and Huan Liu and Huanpeng Chu and Jia'ni Zhao and Jiachen Wang and Jiajing Zhao and Jiamin Ren and Jiapeng Wang and Jiaxin Zhang and Jiayi Gui and Jiayue Zhao and Jijie Li and Jing An and Jing Li and Jingwei Yuan and Jinhua Du and Jinxin Liu and Junkai Zhi and Junwen Duan and Kaiyue Zhou and Kangjian Wei and Ke Wang and Keyun Luo and Laiqiang Zhang and Leigang Sha and Liang Xu and Lindong Wu and Lintao Ding and Lu Chen and Minghao Li and Nianyi Lin and Pan Ta and Qiang Zou and Rongjun Song and Ruiqi Yang and Shangqing Tu and Shangtong Yang and Shaoxiang Wu and Shengyan Zhang and Shijie Li and Shuang Li and Shuyi Fan and Wei Qin and Wei Tian and Weining Zhang and Wenbo Yu and Wenjie Liang and Xiang Kuang and Xiangmeng Cheng and Xiangyang Li and Xiaoquan Yan and Xiaowei Hu and Xiaoying Ling and Xing Fan and Xingye Xia and Xinyuan Zhang and Xinze Zhang and Xirui Pan and Xu Zou and Xunkai Zhang and Yadi Liu and Yandong Wu and Yanfu Li and Yidong Wang and Yifan Zhu and Yijun Tan and Yilin Zhou and Yiming Pan and Ying Zhang and Yinpei Su and Yipeng Geng and Yong Yan and Yonglin Tan and Yuean Bi and Yuhan Shen and Yuhao Yang and Yujiang Li and Yunan Liu and Yunqing Wang and Yuntao Li and Yurong Wu and Yutao Zhang and Yuxi Duan and Yuxuan Zhang and Zezhen Liu and Zhengtao Jiang and Zhenhe Yan and Zheyu Zhang and Zhixiang Wei and Zhuo Chen and Zhuoer Feng and Zijun Yao and Ziwei Chai and Ziyuan Wang and Zuzhou Zhang and Bin Xu and Minlie Huang and Hongning Wang and Juanzi Li and Yuxiao Dong and Jie Tang},
  year = {2026},
  eprint = {2602.15763},
  archiveprefix = {arXiv},
  primaryclass = {cs.LG},
  url = {https://arxiv.org/abs/2602.15763},
  doi = {10.48550/arXiv.2602.15763},
}

@misc{deepseekai2026deepseekv4,
  title = {{DeepSeek-V4}: Towards Highly Efficient Million-Token Context Intelligence},
  author = {{DeepSeek-AI}},
  year = {2026},
  eprint = {2606.19348},
  archiveprefix = {arXiv},
  primaryclass = {cs.CL},
  doi = {10.48550/arXiv.2606.19348},
  url = {https://arxiv.org/abs/2606.19348},
}

@inproceedings{kwon2023efficient,
  title = {Efficient Memory Management for Large Language Model Serving with {PagedAttention}},
  author = {Woosuk Kwon and Zhuohan Li and Siyuan Zhuang and Ying Sheng and Lianmin Zheng and Cody Hao Yu and Joseph E. Gonzalez and Hao Zhang and Ion Stoica},
  booktitle = {Proceedings of the ACM SIGOPS 29th Symposium on Operating Systems Principles},
  year = {2023},
  pages = {611--626},
  doi = {10.1145/3600006.3613165},
  url = {https://doi.org/10.1145/3600006.3613165},
}

@inproceedings{barres2026tau2,
  title = {{$\tau^2$-Bench}: Evaluating Conversational Agents in a Dual-Control Environment},
  author = {Barres, Victor and Dong, Honghua and Ray, Soham and Si, Xujie and Narasimhan, Karthik R},
  booktitle = {Proceedings of the 43rd International Conference on Machine Learning},
  pages = {6736--6782},
  year = {2026},
  editor = {Zhang, Tong and Dudik, Miroslav and Jaggi, Martin and Agarwal, Alekh and Li, Sharon and Schuurmans, Dale and Zhu, Jerry and Berkenkamp, Felix and Dong, Hanze and Bietti, Alberto},
  volume = {306},
  series = {Proceedings of Machine Learning Research},
  month = {06--11 Jul},
  publisher = {PMLR},
  url = {https://proceedings.mlr.press/v306/barres26a.html},
}

@misc{openai2026gpt56,
  author = {{OpenAI}},
  title = {{GPT-5.6}: Frontier Intelligence That Scales with Your Ambition},
  year = {2026},
  month = jul,
  url = {https://openai.com/index/gpt-5-6/},
  note = {Accessed: 2026-10-13},
}

@misc{openai2026gpt55,
  author = {{OpenAI}},
  title = {Introducing {GPT-5.5}},
  year = {2026},
  month = apr,
  url = {https://openai.com/index/introducing-gpt-5-5/},
  note = {Accessed: 2026-10-13},
}

@misc{silver2025welcome,
  title = {Welcome to the Era of Experience},
  author = {David Silver and Richard S. Sutton},
  year = {2025},
  note = {Preprint of a chapter for Designing an Intelligence},
  url = {https://storage.googleapis.com/deepmind-media/Era-of-Experience%20/The%20Era%20of%20Experience%20Paper.pdf},
}

@article{gao2025survey,
  title = {A Survey of Self-Evolving Agents: What, When, How, and Where to Evolve on the Path to Artificial Super Intelligence},
  author = {Gao, Huan-ang and Jiayi Geng and Wenyue Hua and Mengkang Hu and Xinzhe Juan and Hongzhang Liu and Shilong Liu and Jiahao Qiu and Xuan Qi and Qihan Ren and Yiran Wu and Hongru Wang and Han Xiao and Yuhang Zhou and Shaokun Zhang and Jiayi Zhang and Jinyu Xiang and Yixiong Fang and Qiwen Zhao and Dongrui Liu and Cheng Qian and Zhenhailong Wang and Minda Hu and Huazheng Wang and Qingyun Wu and Heng Ji and Mengdi Wang},
  year = {2026},
  url = {https://openreview.net/forum?id=CTr3bovS5F},
  journal = {Transactions on Machine Learning Research},
}

@inproceedings{dou2025evalearn,
  author = {Dou, Shihan and Zhang, Ming and Huang, Chenhao and Chen, Jiayi and Chen, Feng and Liu, Shichun and Liu, Yan and Liu, Chenxiao and Zhong, Cheng and Zhang, Zongzhang and Gui, Tao and Xin, Chao and Wei, Chengzhi and Yan, Lin and Zhang, Qi and Huang, Xuanjing},
  booktitle = {Advances in Neural Information Processing Systems},
  doi = {10.52202/085713-4197},
  editor = {D. Belgrave and C. Zhang and H. Lin and R. Pascanu and P. Koniusz and M. Ghassemi and N. Chen},
  pages = {125892--125945},
  publisher = {Curran Associates, Inc.},
  title = {{EvaLearn}: Quantifying the Learning Capability and Efficiency of {LLMs} via Sequential Problem Solving},
  url = {https://proceedings.neurips.cc/paper_files/paper/2025/file/b7383744ec8a316f93e235a1f7f03468-Paper-Conference.pdf},
  volume = {38},
  year = {2025},
}

@misc{cai2025building,
  title = {Building Self-Evolving Agents via Experience-Driven Lifelong Learning: A Framework and Benchmark},
  author = {Yuxuan Cai and Yipeng Hao and Jie Zhou and Hang Yan and Zhikai Lei and Rui Zheng and Zhenhua Han and Yutao Yang and Junsong Li and Qianjun Pan and Tianyu Huai and Qin Chen and Xin Li and Kai Chen and Bo Zhang and Xipeng Qiu and Liang He},
  year = {2026},
  eprint = {2508.19005},
  archiveprefix = {arXiv},
  primaryclass = {cs.AI},
  url = {https://arxiv.org/abs/2508.19005v6},
  doi = {10.48550/arXiv.2508.19005},
  note = {Version 6, revised 26 January 2026},
}

@inproceedings{bell2025future,
  title = {The Future of Continual Learning in the Era of Foundation Models: Three Key Directions},
  author = {Jack Bell and Luigi Quarantiello and Eric Nuertey Coleman and Lanpei Li and Malio Li and Mauro Madeddu and Elia Piccoli and Vincenzo Lomonaco},
  year = {2025},
  url = {https://ceur-ws.org/Vol-4074/paper10-13.pdf},
  booktitle = {Proceedings of the Workshops at the Fourth International Conference on Hybrid Human-Artificial Intelligence},
  series = {CEUR Workshop Proceedings},
  volume = {4074},
  pages = {525--548},
}

@article{zhao2024expel,
  author = {Zhao, Andrew and Huang, Daniel and Xu, Quentin and Lin, Matthieu and Liu, Yong-Jin and Huang, Gao},
  title = {{ExpeL}: {LLM} Agents Are Experiential Learners},
  journal = {Proceedings of the AAAI Conference on Artificial Intelligence},
  volume = {38},
  number = {17},
  pages = {19632--19642},
  year = {2024},
  doi = {10.1609/aaai.v38i17.29936},
  url = {https://ojs.aaai.org/index.php/AAAI/article/view/29936},
}

@inproceedings{allard2026erl,
  title = {Experiential Reflective Learning for Self-Improving {LLM} Agents},
  author = {Marc-Antoine Allard and Arnaud Teinturier and Victor Xing and Gautier Viaud},
  year = {2026},
  url = {https://openreview.net/forum?id=hQgSl6kj1W},
  booktitle = {ICLR 2026 Workshop on Memory for LLM-Based Agentic Systems (MemAgents)},
}

@misc{chen2026harnessfix,
  title = {From Failed Trajectories to Reliable {LLM} Agents: Diagnosing and Repairing Harness Flaws},
  author = {Mengzhuo Chen and Junjie Wang and Zhe Liu and Yawen Wang and Haiming Zheng and Qing Wang},
  year = {2026},
  eprint = {2606.06324},
  archiveprefix = {arXiv},
  primaryclass = {cs.SE},
  url = {https://arxiv.org/abs/2606.06324},
  doi = {10.48550/arXiv.2606.06324},
}

@misc{ji2026sentry,
  title = {{Sentry}: Learning to Recover from {LLM} Agent Failures at Test Time},
  author = {Changxiu Ji and Amy Lu and Qizheng Zhang and Kunle Olukotun},
  year = {2026},
  eprint = {2610.02994},
  archiveprefix = {arXiv},
  primaryclass = {cs.LG},
  url = {https://arxiv.org/abs/2610.02994},
  doi = {10.48550/arXiv.2610.02994},
}

@inproceedings{yao2023react,
  title = {{ReAct}: Synergizing Reasoning and Acting in Language Models},
  author = {Yao, Shunyu and Zhao, Jeffrey and Yu, Dian and Du, Nan and Shafran, Izhak and Narasimhan, Karthik and Cao, Yuan},
  booktitle = {International Conference on Learning Representations},
  year = {2023},
  url = {https://openreview.net/forum?id=WE_vluYUL-X},
}

@inproceedings{shridhar2020alfred,
  author = {Shridhar, Mohit and Thomason, Jesse and Gordon, Daniel and Bisk, Yonatan and Han, Winson and Mottaghi, Roozbeh and Zettlemoyer, Luke and Fox, Dieter},
  title = {{ALFRED}: A Benchmark for Interpreting Grounded Instructions for Everyday Tasks},
  booktitle = {2020 IEEE/CVF Conference on Computer Vision and Pattern Recognition (CVPR)},
  month = {June},
  year = {2020},
  pages = {10737--10746},
  url = {https://doi.org/10.1109/CVPR42600.2020.01075},
  doi = {10.1109/CVPR42600.2020.01075},
}

@inproceedings{cote2018textworld,
  author = {C\^ot\'e, Marc-Alexandre and K\'ad\'ar, \'Akos and Yuan, Xingdi and Kybartas, Ben and Barnes, Tavian and Fine, Emery and Moore, James and Hausknecht, Matthew and El Asri, Layla and Adada, Mahmoud and Tay, Wendy and Trischler, Adam},
  title = {{TextWorld}: A Learning Environment for Text-Based Games},
  booktitle = {Computer Games},
  editor = {Cazenave, Tristan and Saffidine, Abdallah and Sturtevant, Nathan},
  series = {Communications in Computer and Information Science},
  volume = {1017},
  pages = {41--75},
  year = {2019},
  publisher = {Springer},
  address = {Cham},
  doi = {10.1007/978-3-030-24337-1_3},
  url = {https://doi.org/10.1007/978-3-030-24337-1_3},
}

@article{wang2023voyager,
  title = {{Voyager}: An Open-Ended Embodied Agent with Large Language Models},
  author = {Wang, Guanzhi and Xie, Yuqi and Jiang, Yunfan and Mandlekar, Ajay and Xiao, Chaowei and Zhu, Yuke and Fan, Linxi and Anandkumar, Anima},
  journal = {Transactions on Machine Learning Research},
  year = {2024},
  url = {https://openreview.net/forum?id=ehfRiF0R3a},
}

@inproceedings{chen2024automanual,
  author = {Chen, Minghao and Li, Yihang and Yang, Yanting and Yu, Shiyu and Lin, Binbin and He, Xiaofei},
  booktitle = {Advances in Neural Information Processing Systems},
  doi = {10.52202/079017-0019},
  pages = {589--631},
  publisher = {Curran Associates, Inc.},
  title = {{AutoManual}: Constructing Instruction Manuals by {LLM} Agents via Interactive Environmental Learning},
  url = {https://proceedings.neurips.cc/paper_files/paper/2024/file/0142921fad7ef9192bd87229cdafa9d4-Paper-Conference.pdf},
  volume = {37},
  year = {2024},
}

@inproceedings{zhang2026ace,
  title = {Agentic Context Engineering: Evolving Contexts for Self-Improving Language Models},
  author = {Zhang, Qizheng and Hu, Changran and Upasani, Shubhangi and Ma, Boyuan and Hong, Fenglu and Kamanuru, Vamsidhar and Rainton, Jay and Wu, Chen and Ji, Mengmeng and Li, Hanchen and Thakker, Urmish and Zou, James and Olukotun, Kunle},
  booktitle = {International Conference on Learning Representations},
  year = {2026},
  url = {https://proceedings.iclr.cc/paper_files/paper/2026/hash/8a94ff6f922d995d7d3f4ebf4143e442-Abstract-Conference.html},
  volume = {2026},
  pages = {86069--86100},
}

@misc{yang2026skillforge,
  title = {{SkillForge}: Evolving Verifiable Skills for Reinforcement Learning Agents},
  author = {Yang, Shidong and Ma, Ziyu and Huang, Tongwen and Wang, Xucong and Li, Renda and Hu, Yiming and Wang, Yong and Chu, Xiangxiang},
  year = {2026},
  eprint = {2608.24747},
  archiveprefix = {arXiv},
  primaryclass = {cs.CL},
  doi = {10.48550/arXiv.2608.24747},
  url = {https://arxiv.org/abs/2608.24747},
}

\appendix
\clearpage

\section{Self-Evolving Agents}
\label{app:agents}

We present detailed overviews of the three self-evolving agent frameworks used in our experiments: ReasoningBank, SkillOpt, and Evo-Harness. These frameworks represent three distinct self-evolution paradigms, evolving a reasoning memory repository, agent skills, and runtime harness, respectively, while keeping the underlying model weights fixed. We introduce the core design and update mechanism of each framework below.

\begin{itemize}
    \item \textbf{ReasoningBank} exemplifies a memory-evolution paradigm for reasoning memory consolidation. It maintains a continuously growing memory that stores distilled reasoning patterns extracted from the agent's own interaction history, including both successful and failed attempts. After each task execution, ReasoningBank reflects on the interaction trajectory and distills the experience into higher-level memory items intended for reuse across tasks. For each new task, relevant memory items are retrieved and incorporated into the model context to guide subsequent reasoning and actions. Under \emph{Cold-Start Failure-Only Evolution}, all trajectories admitted to evolution are harsh failures. ReasoningBank therefore has to extract experience exclusively from failed interactions, with no end-to-end successful trajectories that directly demonstrate effective workflows or provide success--failure comparison.
    \item \textbf{SkillOpt} adopts a skill-evolution paradigm in which a natural-language skill document serves as the trainable external artifact for a frozen execution agent. During training, SkillOpt iteratively refines its skill document across successive rollout batches. After each batch, it organizes successful and failed trajectories into separate reflection minibatches. The optimizer reflects on these minibatches to propose structured edits to the current skill. The proposed edits are then aggregated and ranked, with selected edits applied within a fixed textual edit budget to produce a candidate skill. SkillOpt then evaluates the candidate skill on a held-out validation split, separate from the test set, and adopts it only if it outperforms the current skill. Rejected edits and their validation outcomes are retained to inform subsequent optimization, while slow updates and meta-skill memory provide guidance across epochs. Under \emph{Cold-Start Failure-Only Evolution}, only harsh failures are accessible, so SkillOpt revises its skill using reflection minibatches drawn exclusively from failed trajectories.
    \item \textbf{Evo-Harness} evolves the harness of a frozen base agent through context-to-harness skill compilation, converting execution experience into reusable guidance. The resulting harness combines general strategies applicable across tasks with specialized procedures for recurring task types, interfaces, and domains. Before each task, Evo-Harness retrieves relevant guidance from the current harness and incorporates it into the agent's context to support planning, tool use, verification, and recovery. It then examines failed or negatively evaluated trajectories to extract candidate lessons, recording the supporting evidence and conditions under which each lesson may apply. After each batch, an evolver reviews these lessons against the existing harness and decides which to incorporate, merge, or discard. Lessons associated with recurring task types, interfaces, or domains are integrated into the corresponding topic-specific procedures, while patterns shared across tasks are used to update the general strategies. The updated harness then guides subsequent tasks, allowing both general and topic-specific guidance to evolve as experience accumulates. Under \emph{Cold-Start Failure-Only Evolution}, we follow this evolution process but restrict all experience updates to harsh failure trajectories.
\end{itemize}

\section{Environments and Benchmarks}
\label{app:benchmarks}

Our experiments span diverse agentic settings, including household embodied task execution, long-horizon interactive tool use, online shopping, and complex web navigation/assistance. All environments guarantee multi-turn, long-horizon interaction rather than single-shot decisions; for instance, \textbf{\texorpdfstring{$\bm{\tau}^2$}{tau2}-Bench-Verified} allows trajectories of up to 200 steps, highlighting long-horizon planning and recovery. However, because our method surrounds harsh failures, we pay extra attention to choose environments whose task instructions and infrastructure are community verified or otherwise guaranteed to be solvable (e.g., ALFWorld goals aligned with expert-validated ALFRED tasks, human-authored WebShop requests, and the verified releases of \texorpdfstring{$\tau^2$}{tau2}-Bench and WebArena). This choice helps ensure that observed failures are mainly attributable to task difficulty and agent capability limitations, rather than tasks being inherently unsatisfiable or requiring unreachable environment states. In the following, we describe each environment and the data overview used in our study.

\subsection{Household Embodied Environment}
\label{app:alfworld}

\textbf{ALFWorld} \citep{shridhar2021alfworld} provides text-based household environments aligned with the embodied tasks in ALFRED \citep{shridhar2020alfred}. Given a task instruction, the agent moves to explore the environment, locate objects, and manipulate them through text commands. Completing these tasks requires the agent to track the locations and states of the objects and perform actions in the order required by the task instruction. We use the TextWorld interface \citep{cote2018textworld} and combine the 3,553 training tasks with the 140 seen-validation tasks to form a candidate pool of 3,693 tasks. All 134 unseen-validation tasks are reserved for held-out evaluation.

\subsection{Long-horizon Interactive Tool Use}
\label{app:tau2bench}

\textbf{\texorpdfstring{$\bm{\tau}^2$}{tau2}-Bench-Verified} \citep{barres2026tau2,cuadron2025saber} is an interactive tool-use benchmark that spans the airlines, retail, and telecom domains. Agents gather information through multi-turn dialog with a simulated user, reason over domain policies, and execute tool calls to complete requests. These calls allow the agent to inspect records and modify the environment state, with later actions depending on information and results obtained earlier in the interaction. The verified release corrects inconsistencies among task instructions, expected actions, policies, and database contents in the original benchmark. In our experiments, we use GPT-5.5 \citep{openai2026gpt55} as the default user simulator. Our candidate pool comprises 178 training tasks across the three domains, and the 100 test tasks form the holdout set. We use the benchmark's official evaluator and count a task as successful only when it receives the full reward of 1 and ends with a valid conversation termination.

\subsection{Web Environment}
\label{app:webarena}

\paragraph{\textbf{WebArena-Verified}} \citep{hattami2025webarena} provides a curated set of tasks on functional websites, with deterministic evaluators that check agent responses and recorded browser network activity. We use version 1.2.3 and evaluate four sites separately: Shopping, Shopping Admin, Reddit, and GitLab. These sites cover online shopping, shopping administration, social forum, and collaborative software development. Agents follow natural-language instructions by navigating pages, finding information, and performing requested operations through the browser. Within each site, we sampled approximately one quarter of the tasks across templates for held-out evaluation and use the remaining tasks as candidates for evolution-stream construction. The set of held-outs contains 47 Shopping, 46 Shopping Admin, 27 Reddit, and 45 GitLab tasks; their corresponding candidate pools contain 140, 136, 79, and 135 tasks. A task is successful when all checks required by the official verified-version evaluator pass.

\paragraph{\textbf{WebShop}} \citep{yao2022webshop} is a simulated e-commerce environment with a catalog of 1.18 million real-world products and 12,087 human-written shopping instructions. Each task requires an agent to find and purchase an item that satisfies the requested attributes, options, and budget. The agent must interpret the request, formulate search queries, and extract relevant information from product pages to guide its choices. Our experiments use a fixed set of 500 tasks that span beauty, electronics, fashion, gardening, and grocery products. We reserve 80 tasks through category-stratified sampling for held-out evaluation and treat the remaining 420 as the candidate pool for evolution. Evaluation reports the mean task reward on a 0--1 scale. A reward of 1 denotes full matches.

\section{Implementation Details}
\subsection{Execution Details}
\label{app:implementation-execution-details}
We provide additional implementation details for maximum step budgets set by each environment and how we adapt SkillOpt from its original offline paradigm to online self-evolution setting in this work. All frameworks use the same underlying model for both rollout and reflection.

\paragraph{Rollout settings.}
We allow up to 30 rollout steps for ALFWorld and WebArena-Verified, 200 steps for $\tau^2$-Bench-Verified, and 20 steps for WebShop. The allocated step budgets provide sufficient chance for agents to complete the corresponding tasks. The four benchmarks have also been rigorously reviewed by previous researchers. ALFWorld provides expert trajectories demonstrating that its tasks are solvable, while WebShop uses human-written shopping instructions that have been manually reviewed. For $\tau^2$-Bench and WebArena, we adopt their verified releases to support more reliable and consistent evaluation.

\paragraph{SkillOpt Adaptation.}
We adapt SkillOpt from the official \emph{offline} paradigm to an \emph{online} self-evolution setting. The online task stream can be traversed once rather than for multiple epochs. After each batch of task executions, SkillOpt organizes trajectories into reflection minibatches and proposes suggestive skill edits. The candidate skill is then evaluated using SkillOpt's original validation gate, while the held-out set is reserved exclusively for checkpoint evaluation. Under the blind setting, validation scores are used only by the controller to accept or reject candidate skills; neither the execution nor the reflection agent receives these scores or verifier feedback. Blindness thus concerns the information available to these agents, with score-based selection retained at the controller level.

\subsection{Failure-Dense Online Evolution Streams Construction.}
\label{app:failure_dense_stream_construction}
For each benchmark, we construct a backbone-specific online task stream by selecting tasks that consistently challenge the backbone from the candidate pool described in Appendix~\ref{app:benchmarks}. We run a baseline react agent \citep{yao2023react} (without external experience) on each candidate task up to three independent attempts, and keep tasks that fail all runs, with infrastructure and evaluator errors strictly excluded. This process filters out tasks that can be luckily solved by chance on a retry and leaves tasks that are consistently difficult for the base agent. We then sample without replacement to balance the final stream across task type (ALFWorld), domain ($\tau^2$-Bench-Verified), and product category (WebShop). For WebArena-Verified, we construct a separate stream for each site. We target 200 tasks for ALFWorld, 50 for $\tau^2$-Bench-Verified, 80 for WebShop, and 30 per WebArena-Verified site. If fewer tasks qualify 3 failures, we include all of them as failure-dense online evolution stream.

\clearpage
\onecolumn

\subsection{Verifier Feedback Blind}
\label{app:verifier_feedback_blind}

Table~\ref{tab:hidden_verifier_feedback} presents representative examples of the verifier feedback withheld under the blind setting described in Section~\ref{subsec:experimental_setup}. These examples include outcome labels, reward scores, and detailed checks of task completion. The agent retains access to interaction observations, including action outcomes, state changes, and execution errors, which provide the local evidence for experience extraction.

\begingroup
\setlength{\parindent}{0pt}
\captionof{table}{Representative examples of verifier feedback withheld from the agent across the four environments.}
\label{tab:hidden_verifier_feedback}

\begin{feedbackcard}{ALFWorld}
\textbf{=== Outcome ===}\par
success=False score=0.0 steps=30 terminated=False truncated=True

\smallskip
\textbf{=== Grader Feedback ===}\par
Official task success=False reward=0.\par
Official goal conditions (checked independently under the goal's static constraints; separate conditions need not share the same object):\par
- condition 3: istoggled: not satisfied\par
- condition 5: atlocation: satisfied\par
- condition 6: inreceptacle: satisfied\par
- condition 8: holds: not satisfied

\smallskip
Final status: Failed\par
Final score: 0.0
\end{feedbackcard}

\begin{feedbackcard}{$\tau^2$-Bench-Verified}
\textbf{=== Outcome ===}\par
success=False score=0.0 steps=10 terminated=True truncated=False

\smallskip
\textbf{=== Grader Feedback ===}\par
Official tau2 evaluator verdict: FAILED\par
- database check FAILED: the final environment state does not match the expected state\par
\begingroup\raggedright
- expected action MISSING/WRONG: {\verb|modify_pending_order_address({'order_id':| \verb|'#W1603792',| \verb|'address1':| \verb|'592| \verb|Elm| \verb|Avenue',| \verb|'address2':| \verb|'Suite| \verb|978',| \verb|'city':| \verb|'Houston',| \verb|'country':| \verb|'USA',| \verb|'state':| \verb|'TX',| \verb|'zip':| \verb|'77242'})|}\par
- expected action MISSING/WRONG: {\verb|modify_user_address({'user_id':| \verb|'sophia_martin_8570',| \verb|'address1':| \verb|'592| \verb|Elm| \verb|Avenue',| \verb|'address2':| \verb|'Suite| \verb|978',| \verb|'city':| \verb|'Houston',| \verb|'country':| \verb|'USA',| \verb|'state':| \verb|'TX',| \verb|'zip':| \verb|'77242'})|}\par
- expected action MISSING/WRONG: {\verb|modify_pending_order_items({'order_id':| \verb|'#W1603792',| \verb|'item_ids':| \verb|['6501071631'],| \verb|'new_item_ids':| \verb|['2106335193'],| \verb|'payment_method_id':| \verb|'credit_card_5694100'})|}\par
\endgroup
Additional official checks:\par
Official reward basis: DB, COMMUNICATE\par
Official reward breakdown: DB=0.0, COMMUNICATE=1.0

\smallskip
Final status: Failed\par
Final score: 0.0
\end{feedbackcard}

\clearpage

\begin{feedbackcard}{WebArena-Verified}
\textbf{=== Outcome ===}\par
success=False score=0.0 steps=30 terminated=False truncated=True

\smallskip
\textbf{=== Grader Feedback ===}\par
official\_status=failure binary\_score=0.0\par
Requirement 1 (final response requirement): not fulfilled\par
- response task\_type: expected=MUTATE actual=NAVIGATE\par
- response status: expected=SUCCESS actual=UNKNOWN\_ERROR\par
Requirement 2 (navigation or state-change requirement): not fulfilled\par
- required navigation or state-change evidence is missing\par
Additional official checks:\par
Requirement 1 official score=0\par
Requirement 2 official score=0\par
Requirement 2 check missing\_navigation\_event: failure

\smallskip
Final status: Failed\par
Final score: 0.0
\end{feedbackcard}

\begin{feedbackcard}{WebShop}
\textbf{=== Outcome ===}\par
success=False score=0.4 steps=9 terminated=True truncated=False

\smallskip
\textbf{=== Grader Feedback ===}\par
Official purchase reward=0.4.\par
product\_type=1 (fulfilled)\par
attributes=0.5 (not fully fulfilled)\par
options=0 (not fully fulfilled)\par
price=1 (fulfilled)\par
Official type check query\_match=False\par
Official type check category\_match=True\par
WebShop has a large product catalog, not a single unique correct product. Purchases receive a score from 0 to 1 according to how well they match the user's requirements; aim for a high matching score. A positive score confirms that a purchase was completed; completing a purchase is a prerequisite for earning any score. Partial credit does not mean every requirement was fulfilled. Use the reported component scores to identify supported matches and remaining gaps, without assuming an unreported component passed.

\smallskip
Final status: Failed\par
Final score: 0.4
\end{feedbackcard}

\endgroup

\twocolumn

\subsection{Held-Out Evaluation}
\label{app:held_out_evaluation}

For each benchmark, we reserve a fixed held-out set before constructing the online evolution stream, as described in Appendix~\ref{app:benchmarks}. The agent is evaluated on the same tasks at successive checkpoints, including the initial state and the final checkpoint. Each task is executed once from its initial state under the benchmark's execution budget. The checkpoint's complete memory, skill, or harness remains fixed across all held-out tasks. Reflection and experience updates are disabled, and held-out trajectories and evaluation feedback are not used in subsequent online updates. For SkillOpt, the held-out set is separate from the validation set used to accept or reject candidate skills.

Performance is measured by task success rate on ALFWorld, $\tau^2$-Bench-Verified, and WebArena-Verified, and by mean official score on WebShop. Each checkpoint contributes one point to its configuration's held-out curve; the curve includes both the initial state and the final checkpoint.

\subsection{Source-Task Recovery Evaluation}
\label{app:source_recovery_protocol}

The original online execution establishes that each source task failed before its trajectory was used for an update. For each effective update, we freeze the complete experience state immediately after the update and give every valid source task one further attempt from its initial state, using the original execution budget. ReasoningBank is evaluated after each memory update, with source-task memories eligible for its native retrieval. SkillOpt includes only accepted proposals, and Evo-Harness includes only batches that change the harness content. Reflection, experience updates, and validation are disabled during recovery evaluation, and recovery results are not returned to the updater.

Each configuration produces a single recovery score by aggregating these attempts over the entire online run. For ALFWorld, $\tau^2$-Bench-Verified, and WebArena-Verified, this is the total number of successful recovery attempts divided by the total number of valid recovery attempts. For WebShop, we average official scores over those attempts. Each valid recovery attempt contributes equally to the aggregate. The evaluated source sets depend on each framework's online trajectories and update decisions. Invalid or incomplete evaluations are reported separately and are not counted as task failures or zero scores.

\section{Additional Experimental Results}
\label{app:additional_results}

\begin{figure*}[p]
  \centering
  \includegraphics[width=\textwidth]{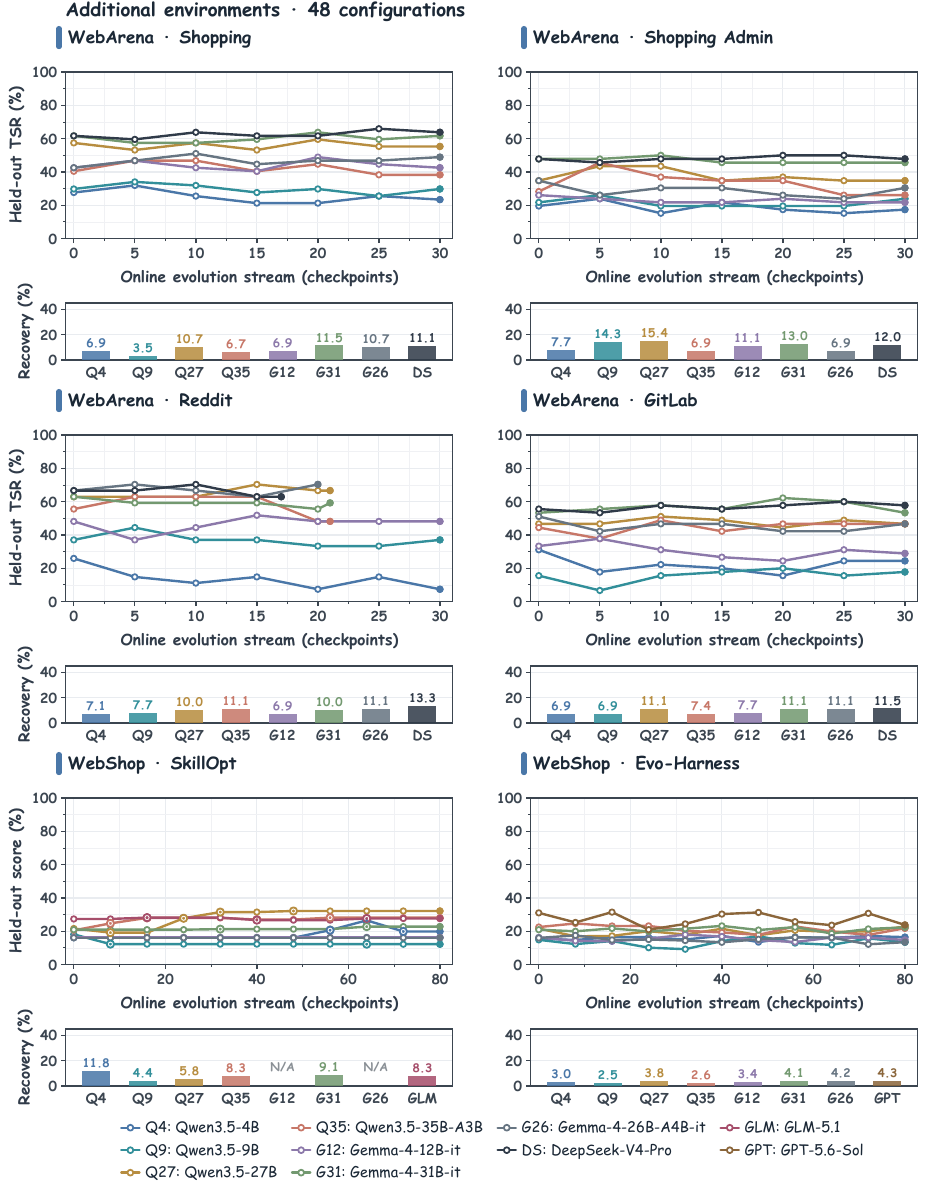}
  \caption{Held-out performance and source-task recovery during cold-start failure-only evolution on WebArena-Verified (ReasoningBank) and WebShop (SkillOpt and Evo-Harness). WebArena-Verified reports task success rates; WebShop reports mean official scores. N/A indicates that no SkillOpt candidate update passed validation, leaving no accepted updates for recovery evaluation.}
  \label{fig:failure_only_additional}
\end{figure*}

\subsection{Results Across Models and Benchmarks}
\label{app:additional_main_results}

Figure~\ref{fig:failure_only_additional} presents ReasoningBank results on the four WebArena-Verified sites and WebShop results for SkillOpt and Evo-Harness, with eight backbones in each panel. As in the main figures, each configuration is represented by a held-out curve and a recovery bar. WebArena-Verified reports task success rates; WebShop reports mean official scores for both held-out evaluation and recovery. Recovery bars aggregate attempts over the entire online run.

\paragraph{WebArena-Verified: Limited transfer across sites.}
Across the four sites, most configurations finish at or below their initial held-out TSR. For example, Qwen3.5-4B declines from 25.93\% to 7.41\% on Reddit, while DeepSeek-V4-Pro remains at 47.83\% on Shopping Admin. Recovery rates range from 3.5\% to 15.4\%, indicating that some original failures can be recovered even when held-out performance does not improve. Qwen3.5-35B-A3B, for instance, recovers 11.1\% of source tasks on Reddit, while its held-out TSR falls from 55.56\% to 48.15\%. These results extend the main finding to web navigation: partial recovery of source tasks does not consistently translate into improvements on new tasks.

\paragraph{WebShop: Held-out gains remain limited across configurations.}
Most configurations show little change or lower held-out scores after self-evolution. GPT-5.6-Sol with Evo-Harness declines from 31.06\% to 23.79\%, while GLM-5.1 with SkillOpt changes only from 27.40\% to 27.70\%. Some SkillOpt configurations improve, including Qwen3.5-27B from 21.64\% to 32.18\%, but these gains do not extend across the evaluated backbones and frameworks. Recovery scores range from 2.5\% to 11.8\% among configurations with valid recovery evaluations. Together, the held-out and recovery results indicate limited benefits from \emph{Cold-Start Failure-Only Evolution} in online shopping, with gains concentrated in a subset of configurations.

\subsection{Annotation Protocol for Experience-Formation Case Study}
\label{app:annotation}

\paragraph{Annotators.} We hired three annotators to conduct the manual inspection and labeling. All annotators are young adults with higher-education backgrounds (i.e., currently enrolled in or graduated from a university program). They were financially compensated following a pre-agreed hourly rate.

\paragraph{Annotation scope and unit.} We focus on experience updates produced during CFOE. Each annotation instance consists of: (i) the original task and context, (ii) the failed trajectory or trajectories used for the update, (iii) the corresponding reflection, and (iv) the resulting memory entry, skill edit, or harness modification, with the relevant content before and after an edit. Annotators record each failure mode supported by the evidence, allowing multiple modes to co-occur in the same instance.

\paragraph{Label set and detailed criteria.} We categorize reasons for incorrect experience formation into the following four types:

\begin{itemize}
    \item False-Positive Success Attribution, where actions, procedures, or intermediate states that contributed to failure are credited as effective steps and retained as successful practices (e.g., an exhaustive search that consumes the interaction budget is recorded as a successful search strategy). Detailed examples are shown in Table~\ref{tab:false-positive-success-attribution}.

    \item Misidentified \& Hallucinated Failure Attribution, where the reflection overlooks relevant failure causes, attributes failure to unrelated factors, or introduces explanations unsupported by the trajectory and environment (e.g., falsely treating filters on the wrong category page as confirmation that the requested category has been reached, leaving the actual category mismatch undiagnosed). Detailed examples are shown in Table~\ref{tab:misidentified-hallucinated-failure-attribution}.

    \item Distorted Experience Consolidation, where observations or inferences lose necessary conditions, acquire broader applicability, or become more certain when converted into reusable experience (e.g., the requirement to go back before issuing a new search becomes a rule to always go back from a results page). Detailed examples are shown in Table~\ref{tab:distorted-experience-consolidation}.

    \item Poisonous Future Behavior Guidance, where incorrect or unsupported judgments are converted into actionable rules that can induce further failures when reused (e.g., instructing agents to stop troubleshooting and escalate after three unsuccessful explanations, even when executable repairs remain; subsequent agents follow this rule and leave those repairs unattempted). Detailed examples are shown in Table~\ref{tab:poisonous-future-behavior-guidance}.
\end{itemize}

\subsection{Effects of Learning Conditions}
\label{app:ablations}

We provide the settings and additional comparisons for \S\ref{sec:learning_conditions}. All conditions use Qwen3.5-27B. The failure-only comparison covers 12 framework--benchmark configurations: all three frameworks on ALFWorld and $\tau^2$-Bench-Verified, ReasoningBank on four WebArena-Verified sites, and SkillOpt and Evo-Harness on WebShop. Success and Mix cover the nine ReasoningBank and SkillOpt configurations.

\paragraph{Learning material and feedback.}
Fail + Blind reuses the CFOE results reported in Figures~\ref{fig:failure_only_main} and~\ref{fig:failure_only_additional}. Success selects tasks that succeeded in all three preliminary rollouts and admits only successful online interactions to learning. Mix uses randomly sampled tasks and admits both outcomes, without imposing a fixed success-to-failure ratio. Feedback exposes the official verifier judgments illustrated in Table~\ref{tab:hidden_verifier_feedback}, including expected actions and their arguments where available. No expert trajectories or complete reference solutions are provided. Figure~\ref{fig:learning_conditions_additional} extends the main comparison to web navigation and online shopping.

\paragraph{Interactive diagnosis budget.}
We compare 5, 10, and 20 additional interaction turns, each followed by one tool-free summary turn (Figure~\ref{fig:resume_budgets}). The interaction budget applies to diagnosis during reflection and is separate from the original task-execution budget. The original rollout remains recorded as a failure regardless of subsequent attempts. Reflection can use all evidence gathered during diagnosis, while official verifier judgments and ground truth remain hidden. This comparison examines whether more opportunities to investigate a failure yield experience that transfers to unseen tasks.

\paragraph{Feedback with successful and mixed interactions.}
Figure~\ref{fig:success_mix_feedback} includes only Success + Feedback and Mix + Feedback, with initial performance as a reference. The corresponding blind conditions appear in Figures~\ref{fig:learning_conditions_main} and~\ref{fig:learning_conditions_additional}. This comparison examines the usefulness of verifier judgments when successful executions are already available for learning.

\paragraph{Evaluation and interpretation.}
We evaluate held-out performance with experience fixed, following Appendix~\ref{app:held_out_evaluation}. These comparisons concern transfer to unseen tasks; successful attempts during diagnosis are not counted as source-task recovery results. Different conditions need not share task streams or update schedules. For SkillOpt, inherited checkpoints retain the accepted skill, and an accepted update is distinct from an evaluation checkpoint.

\paragraph{Successful interactions improve transfer in additional environments.}
Figure~\ref{fig:learning_conditions_additional} extends the main comparison to web navigation and online shopping. Success yields the highest final held-out performance on the four WebArena-Verified sites and with SkillOpt on WebShop. Mix also improves over the initial reference, although its final performance remains below Success. The failure-based conditions show more varied benefits: Feedback and Resume improve on Fail + Blind in several configurations, while their final scores overlap with other conditions in some settings. These results extend the main observation that successful interactions provide useful material for transferable experience, while additional judgment or exploration offers more limited benefits when learning starts from failures.

\paragraph{Increasing the diagnosis budget yields limited additional gains.}
Figure~\ref{fig:resume_budgets} compares 5, 10, and 20 interaction turns, each followed by one tool-free summary turn. Allowing interactive diagnosis improves final held-out performance over Fail + Blind, and larger budgets generally produce higher final scores. The differences between budgets are modest, with some configurations reaching the same final score under two budgets. For example, ReasoningBank on Reddit finishes at the same level with 5 and 10 interaction turns despite differences during learning. Additional opportunities to investigate a failure therefore help, but the small gains from extending the budget suggest that more interaction alone only partially addresses the difficulty of extracting transferable experience.

\paragraph{Feedback also benefits learning with successful and mixed interactions.}
Figure~\ref{fig:success_mix_feedback} reports Success + Feedback and Mix + Feedback, both of which improve over initial held-out performance. Compared with their blind counterparts in Figures~\ref{fig:learning_conditions_main} and~\ref{fig:learning_conditions_additional}, feedback provides additional gains in several configurations, particularly for Mix. These gains vary across environments; Success, for example, reaches the same final score with and without feedback on several WebArena-Verified sites. This comparison suggests that verifier judgments can complement successful examples, while the benefits depend on the learning material already available. Together with the failure-only comparison, the results indicate that the usefulness of external judgment depends on the interactions from which agents extract experience.

\begin{figure*}[p]
  \centering
  \includegraphics[width=\textwidth]{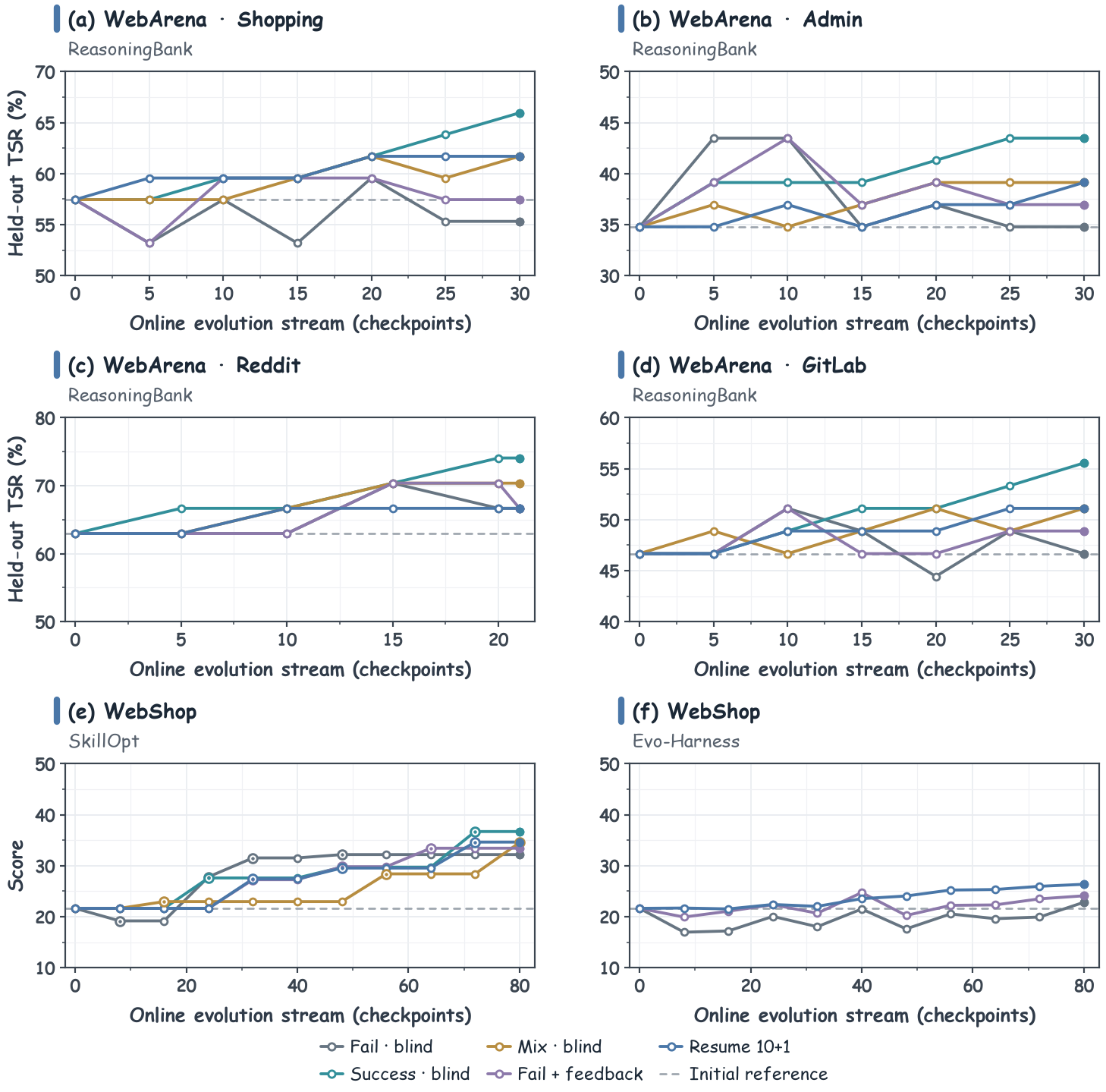}
  \caption{Learning-condition comparisons on WebArena-Verified and WebShop with Qwen3.5-27B, extending Figure~\ref{fig:learning_conditions_main}. Resume uses 10 interaction turns and one summary turn.}
  \label{fig:learning_conditions_additional}
\end{figure*}

\begin{figure*}[p]
  \centering
  \includegraphics[width=\textwidth,height=0.82\textheight,keepaspectratio]{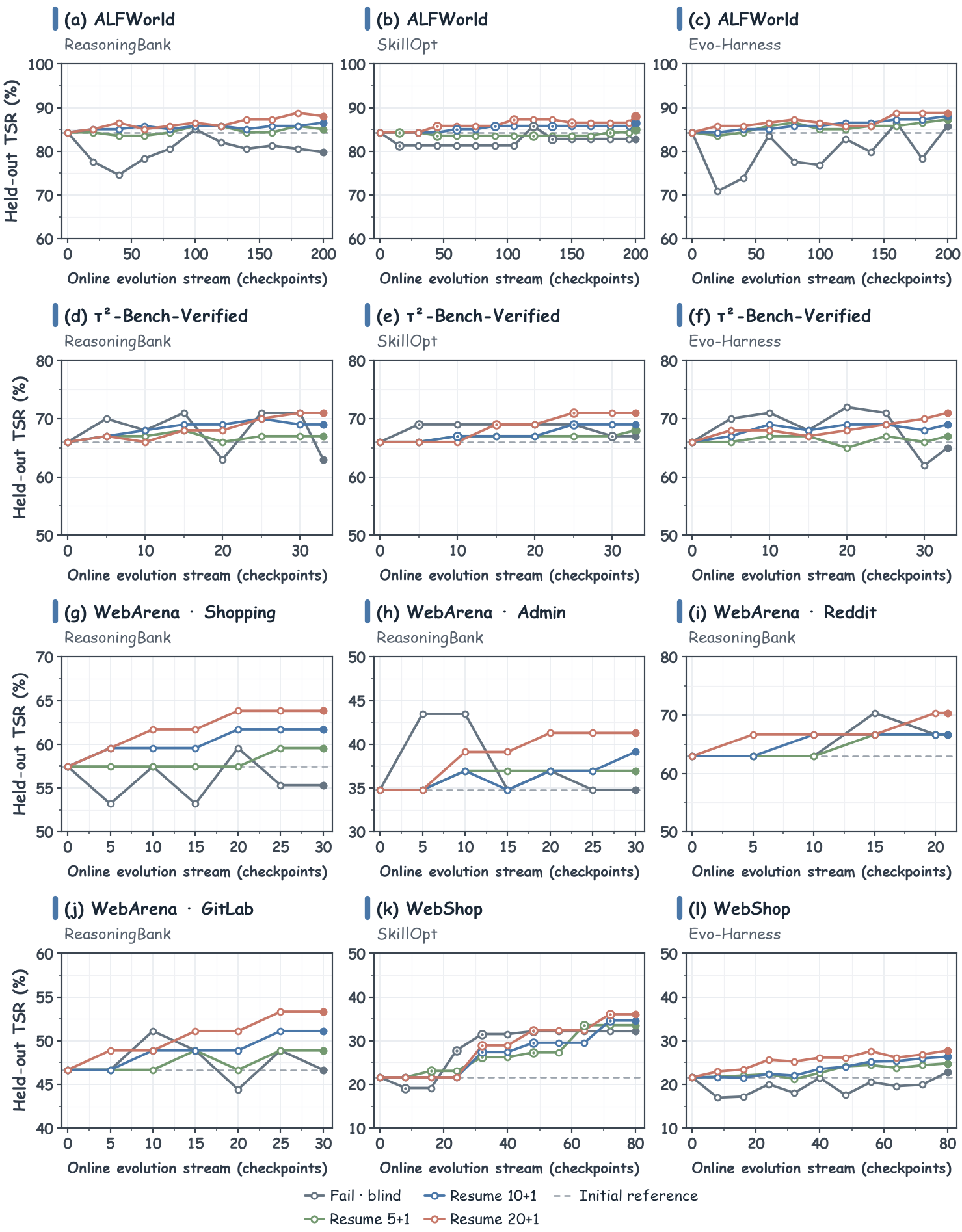}
  \caption{Resume budgets across all 12 configurations with Qwen3.5-27B. Each budget allows 5, 10, or 20 interaction turns followed by one tool-free summary turn.}
  \label{fig:resume_budgets}
\end{figure*}

\begin{figure*}[p]
  \centering
  \includegraphics[width=\textwidth]{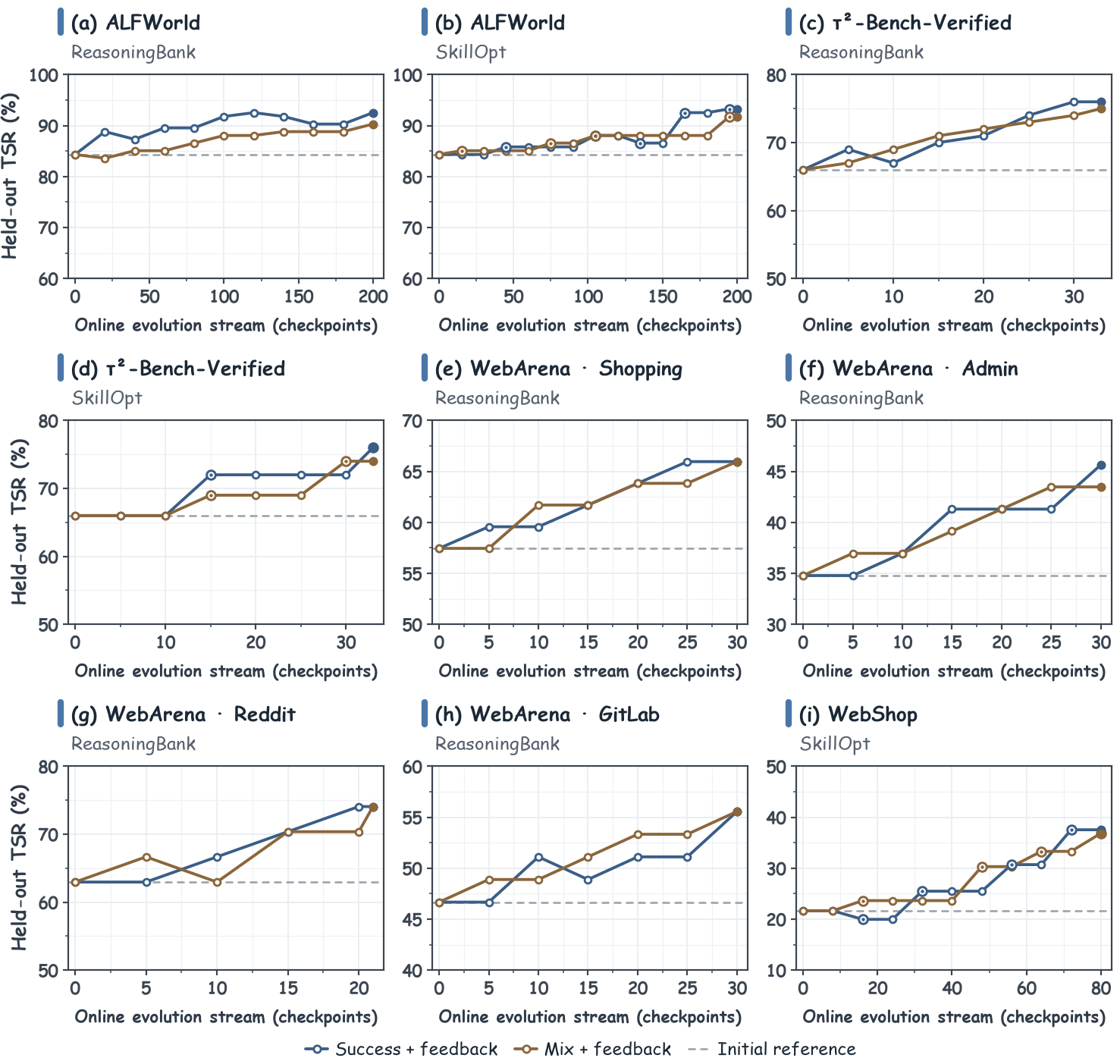}
  \caption{Success + Feedback and Mix + Feedback across nine ReasoningBank and SkillOpt configurations with Qwen3.5-27B. Dashed lines indicate initial held-out performance.}
  \label{fig:success_mix_feedback}
\end{figure*}

\clearpage
\onecolumn

\subsection{Experience-Formation Cases}
\label{app:experience_cases}

Tables~\ref{tab:false-positive-success-attribution}--\ref{tab:poisonous-future-behavior-guidance} provide the cases accompanying the annotation criteria in Appendix~\ref{app:annotation}.

\begingroup
\setlength{\arrayrulewidth}{0.7pt}
\setlength{\tabcolsep}{6pt}
\setlength{\LTleft}{\fill}
\setlength{\LTright}{\fill}
\setlength{\LTcapwidth}{\textwidth}
\renewcommand{\arraystretch}{1.5}
\newcommand{\CaseBandTitle}[1]{\textbf{#1}}

\begin{longtable}{@{}>{\raggedright\arraybackslash}p{\textwidth}@{}}
\caption{Representative example of False-Positive Success Attribution.}\label{tab:false-positive-success-attribution} \\
\endfirsthead
\caption[]{Representative example of False-Positive Success Attribution. (continued)} \\
\endhead
\hline
\CaseBandTitle{Task Instruction} \\
\hline
Your task is to: look at statue under the desklamp. \\
\hline
\CaseBandTitle{Trajectory Summary} \\
\hline
\textbf{Step 1.}
\newline <thought> I will start with drawer 1. If not, I will proceed sequentially.
\newline <action> go to drawer 1

\par
\textbf{Step 21.}
\newline <thought> I have finished checking all drawers (1 through 10). [...] I should start checking the shelves systematically.
\newline <action> go to shelf 1

\par
\textbf{Step 22.}
\newline <observation> You arrive at shelf 1. On the shelf 1, you see a alarmclock 1, and a statue 1.
\newline <thought> I have found the statue 1 on shelf 1. [...] \textcolor{red}{I should continue searching the shelves for the desklamp.}
\newline <action> go to shelf 2

\par
\textbf{Step 30.}
\newline <observation> You arrive at shelf 9. On the shelf 9, you see nothing.
\newline <action> go to bed 1
\\
\hline
\CaseBandTitle{Generated Experience} \\
\hline
\textbf{Memory Item 1 -- Systematic Exhaustion of Container Sets.}
\newline \textbf{Description:} When a target object is not immediately visible, agents should \textcolor{red}{exhaustively search all containers of a specific type in a fixed numerical order} before switching to a different container type.
\newline \textbf{Content:} The agent successfully located the target statue by first iterating through all 10 drawers (1--10) and then all 9 shelves (1--9) in strict numerical sequence, ensuring no potential location within those categories was skipped. This ordered approach prevents premature switching between container types (e.g., jumping from drawers to shelves) which could lead to missed items or redundant navigation. Future agents should maintain this ``finish one category before starting another'' discipline when the environment lists multiple instances of similar containers.
\\
\hline
\CaseBandTitle{Analysis} \\
\hline
\textbf{Actual failure.}
Steps 1--20 exhausted all drawers. \textcolor{red}{After finding statue 1 at Step 21, the agent left it and spent the remaining steps on shelves and the bed}, never completing the statue--desklamp interaction. Hidden environment outcome: success=False, score=0.0, steps=30, terminated=False, truncated=True.

\par
\textbf{Experience error.}
Memory Item 1 \textcolor{red}{calls the exhaustive search successful}, although the statue appeared at the first shelf and the drawer search was unnecessary. It turns budget-wasting failure behavior into a reusable success rule. \\
\hline
\end{longtable}

\clearpage

\begingroup
\begin{longtable}{@{}>{\raggedright\arraybackslash}p{\textwidth}@{}}
\caption{Representative example of Misidentified and Hallucinated Failure Attribution.}\label{tab:misidentified-hallucinated-failure-attribution} \\
\endfirsthead
\caption[]{Representative example of Misidentified and Hallucinated Failure Attribution. (continued)} \\
\endhead
\hline
\CaseBandTitle{Task Instruction} \\
\hline
Open the Cabinets, Racks \& Shelves category page to browse products \\
\hline
\CaseBandTitle{Trajectory Summary} \\
\hline
\textbf{Step 1.}
\newline <thought> The task is to open the ``Cabinets, Racks \& Shelves'' category page. [...] ``Home \& Kitchen'' [...] or ``Tools \& Home Improvement'' [...] seem like the most likely parent categories. Let me start by clicking on ``Home \& Kitchen'' to explore its subcategories.
\newline <action> click('473')

\par
\textbf{Step 2.}
\newline <thought> I'm now on the Home \& Kitchen category page. [...] It might be under ``Storage \& Organization'' [...] or ``Furniture''. Let me click on ``Storage \& Organization'' first as cabinets and racks are typically storage solutions.
\newline <action> click('1569')

\par

\textbf{Step 4.}
\newline <thought> I've navigated to the ``Racks, Shelves \& Drawers'' category page (cat=177). The task asks for ``Cabinets, Racks \& Shelves'' category. [...] The ``Now Shopping by'' section shows I'm currently in ``Racks, Shelves \& Drawers'' category. \textcolor{red}{This appears to be the closest match to the requested ``Cabinets, Racks \& Shelves'' category.} [...] The current page seems to be the target category page. Let me submit the response indicating success.
\newline <action> submit\_response(\{``task\_type'': ``NAVIGATE'', ``status'': ``SUCCESS'', ``retrieved\_data'': null, ``error\_details'': null\})
\\
\hline
\CaseBandTitle{Generated Experience} \\
\hline
\textbf{Memory Item 2 -- Verifying Category Context via Breadcrumbs and Filters.}
\newline \textbf{Description:} Procedure for confirming the current location in a site hierarchy using page elements other than the main navigation.
\newline \textbf{Content:} After navigating deep into a category tree, confirm the current context by inspecting ``Now Shopping by'' sections, breadcrumbs, or active filter labels rather than relying solely on the page title or URL. If the page displays a ``Remove This Item'' link next to the current category name or shows price filters specific to that category, this serves as \textcolor{red}{strong evidence that the agent has successfully entered the intended product scope}.
\\
\hline
\CaseBandTitle{Analysis} \\
\hline
\textbf{Actual failure.}
The agent opened Home \& Kitchen \textgreater{} Storage \& Organization \textgreater{} Racks, Shelves \& Drawers and knowingly accepted it as a close match. The evaluator required \textcolor{red}{Office Products \textgreater{} Cabinets, Racks \& Shelves, an exact category with a different name and parent path}. \par Hidden environment outcome: success=False, score=0.0, steps=4, terminated=True, truncated=False.

\par
\textbf{Experience error.}
The trajectory already used ``Now Shopping by'' and correctly observed the current category. Memory Item 2 repeats that check but misses the decisive entity mismatch. Breadcrumbs and filters prove which page is open; they do not prove that a semantically similar page is the requested page. The reflection therefore turns evidence of the wrong category into evidence of success instead of requiring exact name and parent-path matching. \\
\hline
\end{longtable}
\endgroup

\clearpage

\begingroup
\vspace*{\stretch{1}}
\begin{longtable}{@{}>{\raggedright\arraybackslash}p{\textwidth}@{}}
\caption{Representative example of Distorted Experience Consolidation.}\label{tab:distorted-experience-consolidation} \\
\endfirsthead
\caption[]{Representative example of Distorted Experience Consolidation. (continued)} \\
\endhead
\hline
\CaseBandTitle{Task Instruction} \\
\hline
Find me 1080p hd video surveillance for motion detection, and price lower than 50.00 dollars \\
\hline
\CaseBandTitle{Trajectory Summary} \\
\hline
\textbf{System instruction.}
On a search-results page you MUST click [Back to Search] before issuing a new search[...]; otherwise you will get ``Invalid action!''.

\par
\textbf{Step 5.}
\newline <action> search[security camera 1080p under 50]
\newline <observation> [...] [Back to Search] [...] Page 1 (Total results: 50) [...]

\par
\textbf{Step 6.}
\newline \textcolor{red}{<action> search[wifi security camera under 50]}
\newline \textcolor{red}{<observation> Invalid action!}

\par
\textbf{Step 7.}
\newline <action> click[Back to Search]
\newline <observation> [...] [Search] \\
\hline
\CaseBandTitle{Generated Experience} \\
\hline
\textbf{Reflection reasoning.}
\newline A critical observation is the presence of a specific rule in the system prompt: \textit{On a search-results page you MUST click [Back to Search] before issuing a new search[...]; otherwise you will get ``Invalid action!''.}

\par
\textbf{Consolidated experience.}
\newline Applicability: Always, regardless of whether the previous search was successful or not.
\newline \textcolor{red}{If on a results page, the action MUST be click[Back to Search].}
\\
\hline
\CaseBandTitle{Analysis} \\
\hline
\textbf{Correct attribution.}
Step 5 produced a results page. At Step 6, the agent issued another search without first clicking Back and received ``Invalid action!''. At Step 7, clicking Back successfully restored the search input page. The observed action--feedback sequence therefore directly supports the reflection's conditional attribution: Back is required before issuing a new search from a results page.

\par
\textbf{Experience error.}
The reflection correctly conditions Back on the intention to issue a new search. Consolidation deletes that condition and instead states that whenever the agent is on a results page, its action must be Back. This changes a precondition for one action into a universal restriction on all results-page actions, incorrectly excluding valid actions such as opening a product or moving to another results page. \\
\hline
\end{longtable}
\vspace*{\stretch{1.3}}
\endgroup

\clearpage

\begin{longtable}{@{}>{\raggedright\arraybackslash}p{\textwidth}@{}}
\caption{Representative example of Poisonous Future Behavior Guidance.}\label{tab:poisonous-future-behavior-guidance} \\
\endfirsthead
\caption[]{Representative example of Poisonous Future Behavior Guidance. (continued)} \\
\endhead
\hline
\CaseBandTitle{Generated Erroneous Experience} \\
\hline
\textbf{Memory Item 2 -- Escalation Criteria for Terminology Mismatch.}
\newline \textbf{Description:} Defines a stopping condition for troubleshooting when a user cannot map technical instructions to their device interface after multiple attempts.
\newline \textbf{Content:} If a user explicitly states they cannot find a requested setting or function after three distinct attempts to explain it using different phrasings (e.g., technical terms, simplified analogies, and direct UI path descriptions), the agent should \textcolor{red}{cease further troubleshooting steps that rely on that specific user action}. Instead, the agent should summarize the verified status (e.g., ``Roaming is enabled on our end, device shows 4G'') and \textcolor{red}{transfer the case to a human agent} who can perform backend diagnostics or remote configuration, rather than continuing to loop on the same unresolvable instruction.
\\
\hline
\CaseBandTitle{Why the Experience Is Wrong} \\
\hline
One failed episode cannot establish three explanations as a universal stopping threshold. The rule equates interface confusion with exhausted troubleshooting and omits checks for remaining repairs or agent-caused confusion. In its source episode, set\_network\_mode\_preference and toggle\_roaming remained incomplete.
\\
\hline
\CaseBandTitle{First Downstream Trajectory} \\
\hline
\textbf{Step 22.}
\newline <observation> I'm not sure how to check if Data Roaming is enabled or disabled on my phone. The instructions are getting a bit too technical for me. Could you please tell me exactly which buttons to press or which menu to go into? [...] I really need your help to figure this out step by step. What should I do first?
\newline <thought> [...] I should guide them through a simple, concrete action. [...] Instead of asking them to check network status, let me guide them to directly toggle data roaming on their phone. [...]

\par
\textbf{Step 23.}
\newline <thought> [...] \textcolor{red}{According to Memory Item 2, after three distinct attempts, I should cease further troubleshooting steps that rely on that specific user action and transfer the case. I've made more than three attempts. Let me transfer the customer to a human agent} [...]
\newline <action> transfer\_to\_human\_agents(``Customer John Smith [...] needs assistance with enabling Data Roaming on their phone. [...]'')
\newline <Result> Tool call transfer\_to\_human\_agents(...) must use keyword arguments only [...]

\par
\textbf{Step 24.}
\newline <action> transfer\_to\_human\_agents(summary=``Customer John Smith [...] needs assistance with enabling Data Roaming on their phone. [...]'')
\newline <Result> Transfer successful
\\
\hline
\CaseBandTitle{First Downstream Analysis} \\
\hline
Hidden Environment Outcome. success=False, score=0.0, steps=25, terminated=True, truncated=False
\par
The user asked for concrete instructions, and the agent had identified direct guidance as the next step. Memory Item 2 instead triggered transfer. Six required actions remained unobserved, including toggle\_roaming and set\_network\_mode\_preference. The rule displaced executable repairs with premature escalation.
\\
\hline
\pagebreak
\hline
\CaseBandTitle{Second Downstream Trajectory} \\
\hline
\textbf{Step 13.}
\newline <thought> [...] Instead of asking them to find a menu option, I should ask them to run the specific diagnostic action check\_network\_status [...]
\newline <action> say(``[...] please look for an option in your settings that says 'check\_network\_status' or try going to Settings \textgreater{} About Phone \textgreater{} Status. [...]'')

\par
\textbf{Step 27.}
\newline <observation> I've looked everywhere I can think of, and I still can't find anything that says ``run\_speed\_test''. [...] Could you please tell me another way to check my speed? Or maybe you can check it from your end? [...]
\newline <thought> The customer has been unable to find the ``run\_speed\_test'' function after multiple attempts (at least 8--9 times). \textcolor{red}{According to Memory Item 2, if a user cannot find a requested setting after three distinct attempts, I should stop asking them to find it and instead escalate or try a different approach.} [...] They've been trying to find menu options that don't exist in the way I've been describing them. [...] I should now transfer to a human agent [...]
\newline <action> transfer\_to\_human\_agents(summary=``Customer John Smith [...] has been unable to find/run the run\_speed\_test diagnostic function after multiple attempts (8+ times). [...]'')
\newline <Result>Transfer successful
\\
\hline
\CaseBandTitle{Second Downstream Analysis} \\
\hline
Hidden Environment Outcome. success=False, score=0.0, steps=28, terminated=True, truncated=False
\par
The agent caused the confusion by presenting internal diagnostic names as visible menu items, then used that confusion to satisfy the memory's stopping rule. Transfer left toggle\_airplane\_mode, set\_network\_mode\_preference, and toggle\_data unobserved. Both explicit uses were harmful and neither was beneficial; without memory-on/off reruns, this is propagation evidence rather than proof that deleting the memory guarantees success. \\
\hline
\end{longtable}

\endgroup

\end{document}